\documentclass[3p]{elsarticle}

\usepackage{hyperref}

\journal{Transportation Research Part C}

\biboptions{authoryear}

\usepackage{multirow}
\usepackage{graphicx}
\usepackage{subfig}
\usepackage{amsmath}
\usepackage{amssymb}
\usepackage{hyperref}
\usepackage{booktabs,tabularx}
\newcolumntype{x}{>{\centering\arraybackslash}X}
\usepackage{float}

\newcommand{\ie}{\mbox{i.e.} }
\newcommand{\eg}{\mbox{e.g.} }

\newcommand{\transpose}{{\mbox{\scriptsize T}}}

\newcommand{\N}{\mathcal{N}}
\newcommand{\E}{\mathbb{E}}
\newcommand{\bs}{\boldsymbol}
\newcommand{\bx}{\textbf{x}}

\newcommand{\by}{\textbf{y}}

\newcommand{\bV}{\textbf{V}}

\newcommand{\bff}{\textbf{f}}
\newcommand{\bgg}{\textbf{g}}

\newcommand{\bt}{\textbf{t}}

\begin{document}

\begin{frontmatter}

\title{Heteroscedastic Gaussian processes for uncertainty modeling in large-scale crowdsourced traffic data\footnote{DOI: https://doi.org/10.1016/j.trc.2018.08.007\\
URL: https://www.sciencedirect.com/science/article/pii/S0968090X18300147}}


\author[mymainaddress]{Filipe Rodrigues\corref{mycorrespondingauthor}}
\ead[url]{http://fprodrigues.com}

\cortext[mycorrespondingauthor]{Corresponding author}
\ead{rodr@dtu.dk}

\author[mymainaddress,mysecondaryaddress]{Francisco C. Pereira}
\ead{camara@dtu.dk}

\address[mymainaddress]{Technical University of Denmark (DTU), Bygning 116B, 2800 Kgs. Lyngby, Denmark}
\address[mysecondaryaddress]{Massachusetts Institute of Technology (MIT), 77 Mass. Ave., 02139 Cambridge, MA, USA}

\begin{abstract}
Accurately modeling traffic speeds is a fundamental part of efficient intelligent transportation systems. Nowadays, with the widespread deployment of GPS-enabled devices, it has become possible to crowdsource the collection of speed information to road users (\mbox{e.g.} through mobile applications or dedicated in-vehicle devices). Despite its rather wide spatial coverage, crowdsourced speed data also brings very important challenges, such as the highly variable measurement noise in the data due to a variety of driving behaviors and sample sizes. When not properly accounted for, this noise can severely compromise any application that relies on accurate traffic data. 
In this article, we propose the use of heteroscedastic Gaussian processes (HGP) to model the time-varying uncertainty in large-scale crowdsourced traffic data. Furthermore, we develop a HGP conditioned on sample size and traffic regime (SRC-HGP), which makes use of sample size information (probe vehicles per minute) as well as previous observed speeds, in order to more accurately model the uncertainty in observed speeds. Using 6 months of crowdsourced traffic data from Copenhagen, we empirically show that the proposed heteroscedastic models produce significantly better predictive distributions when compared to current state-of-the-art methods for both speed imputation and short-term forecasting tasks. 
\end{abstract}


\end{frontmatter}


\section{Introduction}
\label{sec:introduction}

Modeling traffic speeds is an essential task for developing intelligent transportation systems, because it provides real-time and anticipatory information about the performance of the network. This information is not only essential for traffic managers, since it allows them to properly allocate resources (\eg control traffic lights) and identify problematic situations, but it also helps users to make better travel decisions by providing them with a complete ``picture" of the traffic status throughout the city (e.g., suggest to take an alternative route or delay/advance the departure) \citep{liu2013adaptive}. The role of accurate traffic speed modeling is even more significant when we consider innovative car-sharing, autonomous vehicles and connected vehicles technologies \citep{tajalli2018distributed}, where inappropriate routing of vehicles and poor system-wide optimization and coordination can have severe adverse effects in the behavior of the road network (e.g., congestion and poor quality of service) and, ultimately, it can be decisive to the adoption of these technologies. 

There are two main sources of traffic speed data: static traffic sensors located at fixed location and GPS sensors from floating vehicles. While traditional speed modeling approaches tend to rely solely on static traffic sensors, which are accurate but expensive to deploy and maintain, nowadays, with the development and widespread deployment of GPS-enabled devices, it has become possible to achieve a much better sensing coverage of the entire road network. In fact, the development of crowdsourcing technologies, where individual users contribute with their own GPS data from their mobile devices, further provides a unique potential for obtaining rather accurate, inexpensive and complete measurements of the speed conditions throughout the network. Hence, it is not surprising that this type of traffic data is becoming increasing popular among traffic managers, operators and local authorities, with many of these acquiring traffic data consisting of aggregated speed measurements from probe vehicles from providers such as INRIX\footnote{http://inrix.com} or HERE\footnote{http://here.com}. However, despite its potential, this data also brings many interesting challenges. 

A key fundamental challenge for using crowdsourced speed data in practice is accurately modeling the uncertainty associated with it. Since this data typically consists of aggregated speeds based on individual GPS measurements from a heterogeneous fleet of contributing vehicles and devices (probe vehicles/devices), the resultant speed information can be extremely noisy. This can be due to several reasons, such as low number of samples (probe devices), accuracy of the GPS-enabled devices (as studied by \cite{guido2014treating}), different drivers' behavior, etc. As a consequence, in some situations, the overall picture of the traffic conditions that this data provides can be significantly blurred, causing applications that rely on it to fail. For example, anomaly detection algorithms can be misled to believe that there is something wrong in a certain road segment, when the problem is simply due to momentarily poor data quality. Similarly, forecasting algorithms can be led to produce erroneous predictions by not accounting for the noise in the speed data when modeling it. 


This article proposes the use of heteroscedastic Gaussian processes in order to produce models that account for the non-constant variance of speeds through time. Gaussian processes (GPs) are flexible non-parametric Bayesian models that are widely used for modeling complex time-series. Indeed, GPs have been successfully applied to model and predict with state-of-the-art results various traffic related phenomena such as traffic congestion \citep{liu2013adaptive}, travel times \citep{ide2009travel}, pedestrian and public transport flows \citep{neumann2009stacked,rodrigues2016bayesian}, traffic volumes \citep{xie2010gaussian}, driver velocity profiles \citep{armand2013modelling}, etc. The fully Bayesian non-parametric formulation of GPs makes them particularly well suited for modeling uncertainty and noise in the observations. Heteroscedastic approaches using GPs to model complex noisy time-series additionally extend the capabilities of GPs to capture the uncertainty in the data by allowing the latter to vary between different time periods. In this article, we take these approaches one step further by proposing a heteroscedastic Gaussian process in which the speed variance is conditioned on the observed sample size, \mbox{i.e.} the number of vehicles/devices per minute (which can also be regarded as a noisy proxy for traffic flow), and on the current traffic regime. The intuition is that the uncertainty associated with a speed observation varies with the number of samples (vehicles) that were used to produce that observation, with more samples producing more accurate speed measurements, as well as the traffic regime. As it turns out, conditioning the observation uncertainty on the number of samples per time interval leads to significantly more accurate predictive distributions. 

Using a large-scale dataset of crowdsourced speeds provided by Google for Copenhagen, we consider two major tasks: speed imputation and short-term forecasting. Speed imputation refers to the post-hoc problem of predicting the speeds that were not observed due to the absence of sensing devices traveling along a road segment or due to any other data collection issue. On the other hand, short-term forecasting refers to the problem of predicting the speeds in a road network for short periods ahead of time (typically 5 to 15 minutes). While short-term forecasting is a fundamental part of any intelligent transportation system, speed imputation can be critical for the success of any application that makes use of that type of data, especially when the missing observation rate is higher, as it is common with crowdsourced data.

By applying the proposed heteroscedastic GP model to crowdsourced speed data, we are effectively able to quantify the uncertainty in the observed speeds and obtain significantly more accurate predictive distributions for traffic speed imputation and short-term forecasting. This, in turn, allows us to produce precise prediction intervals, which are of vital importance for many real-world applications. In fact, the value of accurate prediction intervals is often neglected in the transportation literature. However, for various tasks, it is often more important to be able to estimate prediction intervals than single point estimates. For example, when planning a trip, it is common the case where we need to guarantee with some level of confidence (e.g., 95\%) that the users will arrive on time to the destination. Similarly, when modeling travel demand, such as for public transport or autonomous vehicles, it is essential to ensure that the allocated resources are enough to accommodate the demand. Therefore, rather than planning and allocating resources in accordance to the estimated mean demand, it is better to rely on a different quantile of the predictive distribution in order to guarantee quality of service and avoid travelers' dissatisfaction and frustration. However, if the prediction intervals are not accurate enough, we risk either underestimating our uncertainty, causing problems to the users, or overestimating it, thus wasting valuable resources. By providing a heteroscedastic treatment of the speed data, we are able to produce accurate prediction intervals, while also reducing the error of the mean predictions. Furthermore, by using an approximate inference technique based on variational inference, we are also able to scale the proposed approach to relatively large datasets such as the one used in our experiments. 


The rest of this article is organized as follows. First, Section~\ref{sec:related_work} reviews related works. Section~\ref{sec:gp} introduces GPs and discusses how to use them for modeling time-series data. The proposed sample-size-and-regime-conditioned heteroscedastic GP (SSRC-HGP) is presented in Section~\ref{sec:fc-hgp}. 
A thorough experimental evaluation of the proposed methodology in comparison with other state-of-the-art approaches is presented in Section~\ref{sec:experiments}. Finally, we conclude in Section~\ref{sec:conclusion}.

\section{Related work}
\label{sec:related_work}

\subsection{Traffic speed modeling}

Traffic speed modeling is a core component of transportation systems that has been researched for several decades now. Particularly due to wide variety of data sources and modeling objectives, the body of literature dealing with traffic speed data is rather extense. Therefore, this section attempts to provide only an overview of the field, with particular emphasis on approaches that are related to the ideas proposed in this article.

In terms of sources, the traffic speed data considered in the literature is usually obtained from either dynamic traffic measurements obtained from individual GPS sensors or static traffic sensors at fixed locations. The latter typically have the advantage that they are less susceptible to the effects of external factors and therefore tend to produce more reliable measurements of speed, although they are still prone to occasional sensor malfunctions. However, their deployment and maintenance can be very expensive and, therefore, as recent surveys show, their use in major cities is restricted to only a few road segments \citep{schafer2002traffic}. Contrarily, speed data from individual GPS sensors is considerably cheaper to obtain and its collection is not spatially restricted to a few predefined places. 
In this article, we consider this type of traffic speed data. 

Regarding modeling techniques, traditional approaches can be roughly categorized in two main groups: parametric and non-parametric approaches \citep{vlahogianni2004short}. Within the parametric approaches category, a significant part of the literature is dedicated to statistical parametric models such as auto-regressive (AR), auto-regressive moving average (ARMA) models and other variants. These models developed from the time-series community find their strengths in their probabilistic treatment, simplicity and computational efficiency \citep{box2015time}. Therefore, they very early found their way into the area of transportation systems (\eg \cite{davis1991nonparametric,williams2003modeling}) and to this day they are considered standard baseline approaches. Another popular approach for traffic speed modeling is neural networks. Although these powerful non-linear models have been successfully applied for predicting various traffic-related phenomena such as predicting speeds \citep{dia2001object} and flows \citep{polson2017deep}, they are severely undermined by their black-box nature and lack of probabilistic treatment. The models described in this article distinguish themselves from these by following a non-parametric approach to time-series. 

As the name suggests, non-parametric approaches do not assume any specific functional form for the dependent and independent variables. Instead, the idea is to base regression on a notion of similarity. Popular non-parametric approaches to time-series modeling range from simple nearest-neighbors, which were applied to traffic flow in \cite{oswald2001traffic}, to more powerful kernel methods such as support vector regression (SVR). The latter were successfully applied, for example, to predict travel times \citep{wu2004travel}. More recently, with the developments in the Gaussian process (GP) literature, researchers started adopting them for various traffic related problems. Indeed, the fully Bayesian non-parametric formulation of GPs makes them particularly well suited for complex and noisy time-series. Furthermore, the explicit probabilistic interpretation of the GP outputs and their ability to estimate predictive uncertainty makes GPs ideal candidates for modeling traffic phenomena. Therefore, it is not surprising that GPs have been shown to outperform traditional approaches and produce state-of-the-art results for various transportation-related problems such as mobility demand prediction \citep{chen2013gaussian}, traffic speed forecasting \citep{min2011real} and imputation \cite{rodrigues2018multi}, traffic volume prediction \citep{xie2010gaussian}, travel time prediction \citep{ide2009travel} and adaptive vehicle routing in congestion environments \citep{liu2013adaptive}. 

\subsection{Heteroscedastic time-series modeling}

For practical reasons, time-series data is often treated as having more signal than noise, and a well-behaved noise structure. This structure usually relies on assuming non-biased models and with constant variance (homoscedasticity), typically formulated as a white noise Gaussian distribution \citep{antunes2017review}. Unfortunately, it turns out that rather frequently, reality is not as ``well behaved" and such assumptions may become unrealistic and inappropriate. This is especially true if we consider complex traffic behaviors and heterogeneous data sources that rely on noisy sensors such as the one considered in this article. However, with a few notable exceptions such as \cite{tsekeris2009short} and \cite{chen2011short}, who explore the use of GARCH volatility models, and \cite{lin2018quantifying}, who consider the direct estimation of prediction intervals, the heteroscedastic treatment of traffic phenomena, like the one proposed in this article, has been studied to a much smaller extent. 

In the time-series literature, the importance of modeling heteroscedastic time-series has been well recognized. Causes for heteroscedasticity can vary from case to case, but most of them are related to the model misspecification, measurement errors, sub-population heterogeneity, noise level or it is just a natural intrinsic property of the dataset \citep{antunes2017review}. Regardless of the cause, several approaches have been proposed to deal with heteroscedastic time-series, although the majority have been focused on finance data \citep{hamilton1994autoregressive}, where time-dependent volatility takes its upmost form. These include models such as ARCH, ARMA-CH and other variants \citep{gourieroux2012arch,chen2011short}. Due to the recent increase in interest of the statistics and machine learning communities in Gaussian processes, various approaches have been proposed for developing heteroscedastic GP models \citep{kersting2007most,quadrianto2009kernel,titsias2011variational}. The key difficulty there typically lies in finding efficient approximate inference algorithms. While earlier approaches relied on computationally expensive Markov chain Monte Carlo (MCMC) techniques \citep{goldberg1998regression}, it was not until more recently that researchers started developing deterministic techniques such as the a maximum-a-posteriori (MAP) approach proposed in \cite{kersting2007most,quadrianto2009kernel}. However, MAP estimation does not integrate out all latent variables and is prone to overfitting. With that in consideration, \cite{titsias2011variational} proposed a variational approximation that allows for accurate inference in heteroscedastic GPs with a computational cost similar to that of standard GPs. The authors empirically showed that their proposed approach outperforms popular state-of-the-art volatility models such as GARCH in benchmark financial time-series data. In this article, we follow the same variational approximation proposed by \cite{titsias2011variational}. 

Despite the success of heteroscedastic GPs in modeling real-world phenomena such as wind speeds \citep{jiang2010adaptive} or biophysical variables \citep{lazaro2014retrieval}, where they have been shown to produce state-of-the-art results, their application to modeling complex traffic phenomena like traffic speeds has, to the best of our knowledge, never been researched. As we empirically demonstrate, the heteroscedastic treatment of crowdsourced speed data using Gaussian processes not only enables us to achieve better predictive distributions and more accurate prediction intervals for the predictions, but it also allows us to improve over the predictive performance of state-of-the-art methods. In fact, we further propose a heteroscedastic GP conditioned on sample size, which allows us to obtain significantly more accurate predictive distributions. 

\section{Gaussian processes for time-series}
\label{sec:gp}

Let us start by casting the problem of modeling a time-series of $T$ observations, $\by = \{y_1,...,y_T\}$, as a regression problem of the form
\begin{align}
	y_t = f(t) + \epsilon,
	\label{eq:homeskedastic_formulation}
\end{align}
where $f$ is an unknown function of time (and possibly other variables), and $\epsilon$ is typically an additive white noise process such that $\epsilon \sim \N(\epsilon|0,\sigma^2)$. Traditional approaches to time-series, such as autoregressive or ARIMA models, proceed by assuming a parametric linear form for $f$, whose parameters are then typically estimated by exploiting the maximum likelihood principle. In contrast, GP approaches to time-series modeling assume $f$ to be a non-linear non-parametric function and proceed by placing a Gaussian process prior over $f$.

A Gaussian process is defined as a collection of random variables, any finite number of which have (consistent) joint Gaussian distributions \citep{Rasmussen2005}. Let us consider a multivariate (joint) Gaussian distribution, $\N(\bff|\bs\mu,\bs\Sigma)$, over the T-dimensional vector $\bff = (f(t_1),\dotsc,f(t_T))^\transpose$. While a multivariate Gaussian distribution is fully specified by a mean vector $\bs\mu$ and a covariance matrix $\bs\Sigma$, a GP is a stochastic process fully specified by a mean function $m(t) = \E[f(t)]$ and a positive definite covariance function $k(t,t') = \mbox{cov}[f(t),f(t')]$. By making use of the mean and covariance functions, GPs specify a way to determine the mean of any arbitrary point in time $t$ and how that point covaries with the nearby points. We can then think of GPs as a generalization of a multivariate Gaussian distribution to infinitely many variables. If we loosely see a function as a infinitely long vector $\bff$, where each entry specifies the function value $f(t)$ for a particular time $t$, then we can see a GP as a probability distribution over functions. 

\begin{figure}
	\centering
	\subfloat[squared exponential (SE)]{\includegraphics[width=0.35\linewidth]{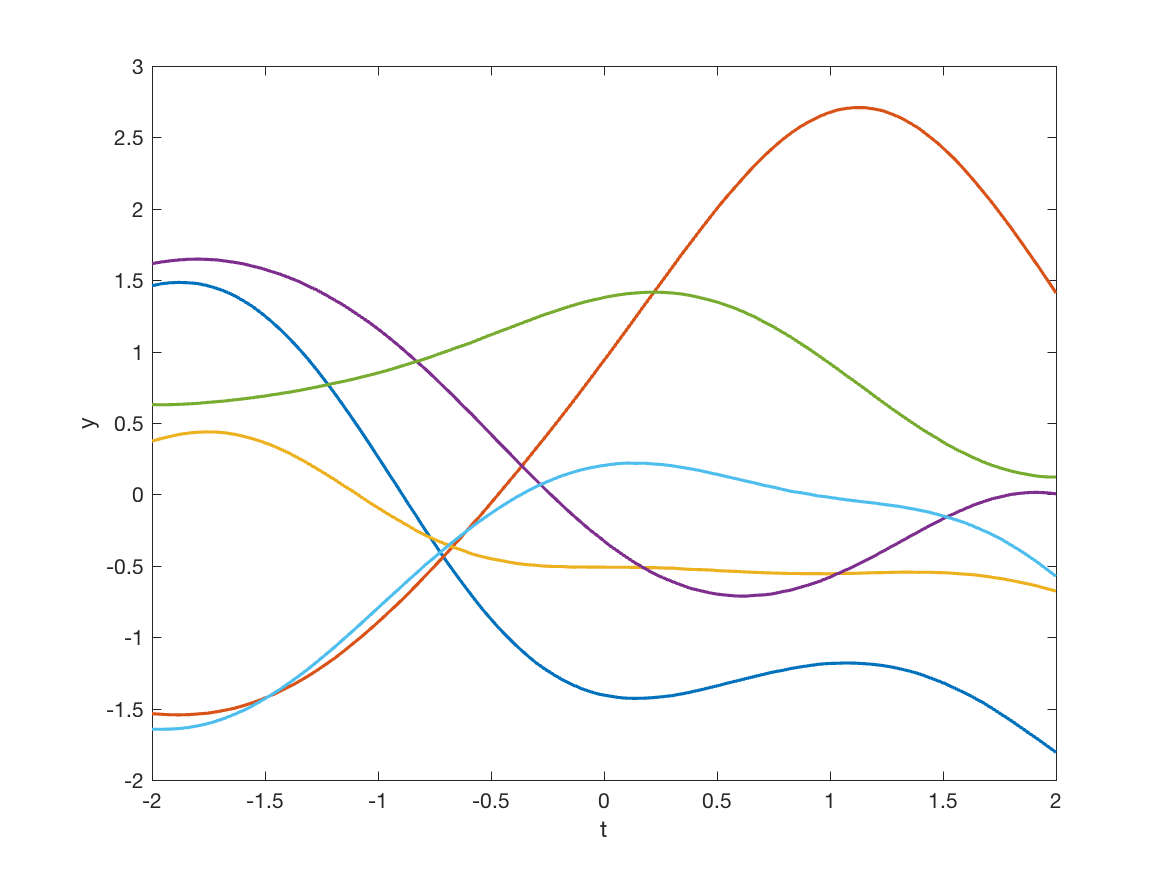}\label{fig:gp_sample_SE}}\hspace{-0.2cm}
	\subfloat[periodic (PER)]{\includegraphics[width=0.35\linewidth]{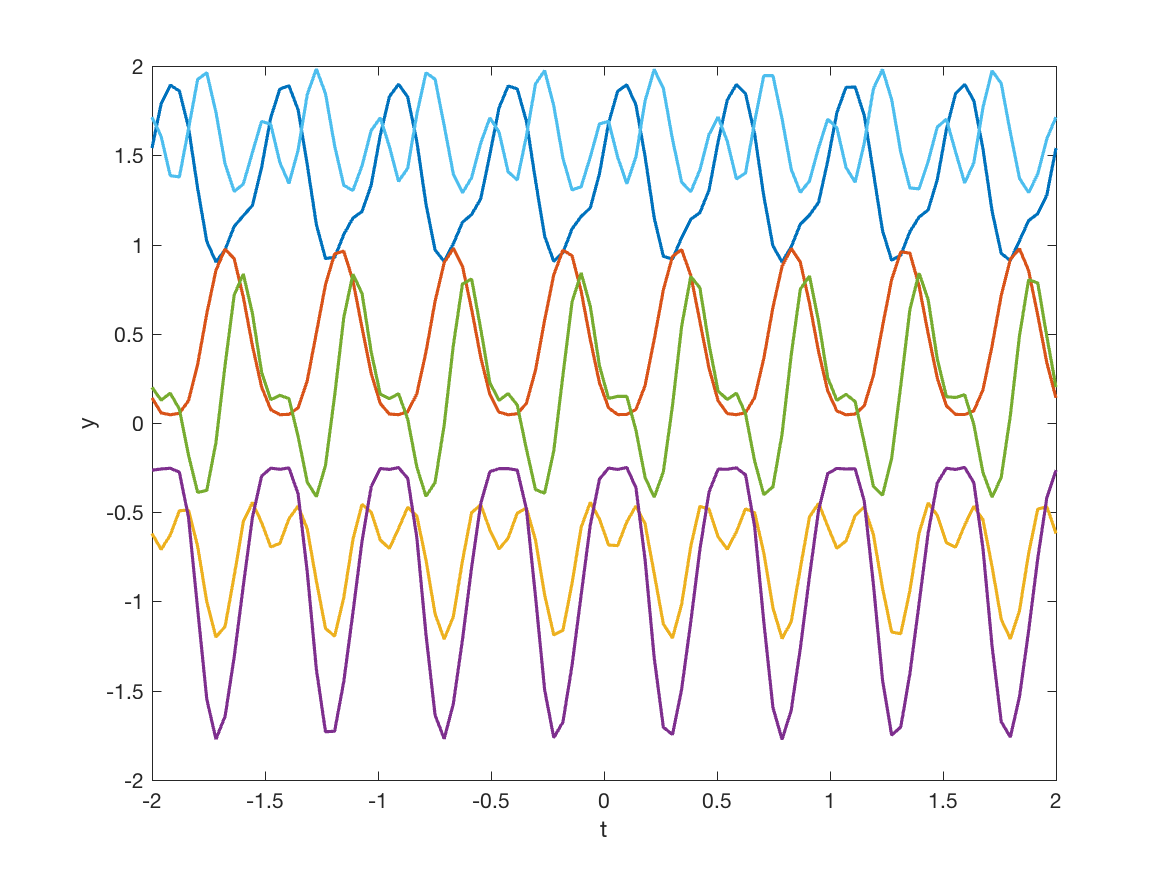}\label{fig:gp_sample_PER}}
	
	\vspace{-0.35cm}
	\subfloat[white noise (WN)]{\includegraphics[width=0.35\linewidth]{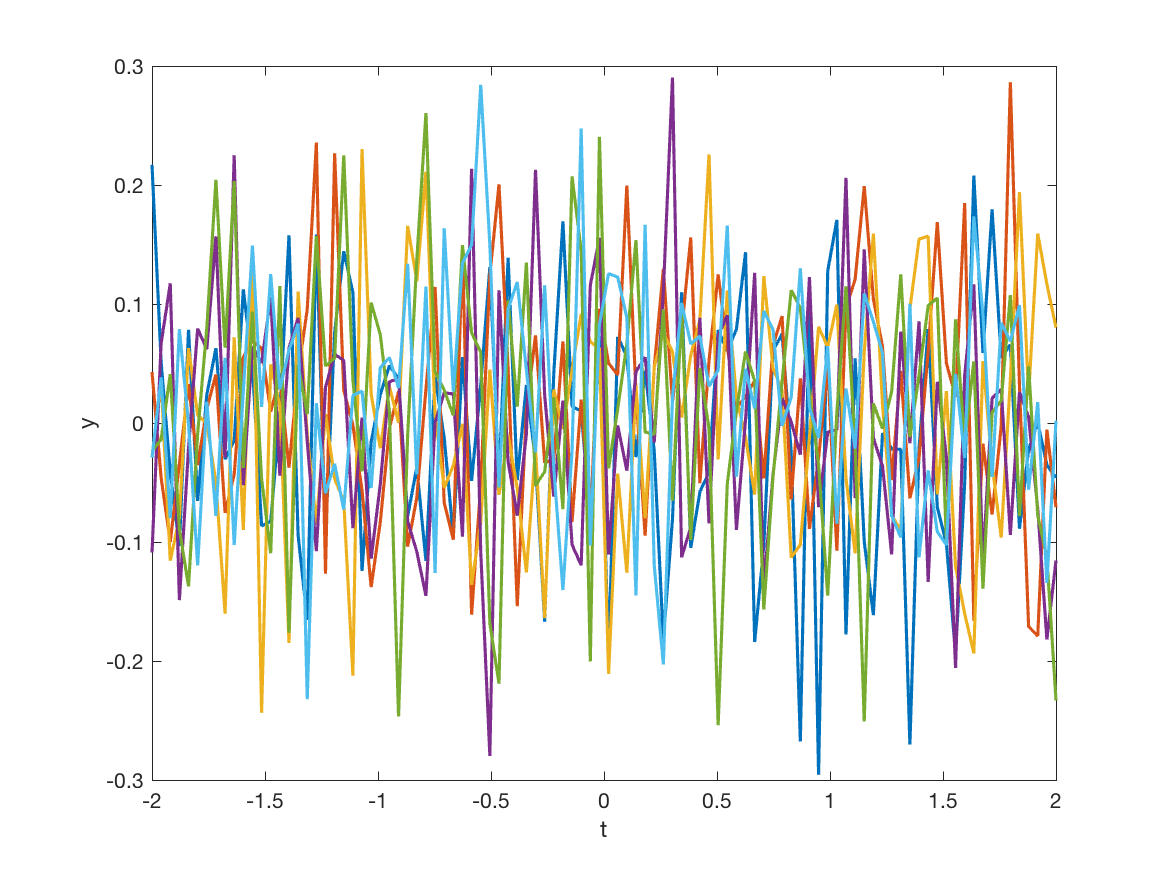}\label{fig:gp_sample_WN}}\hspace{-0.2cm}
	\subfloat[SE+WN]{\includegraphics[width=0.35\linewidth]{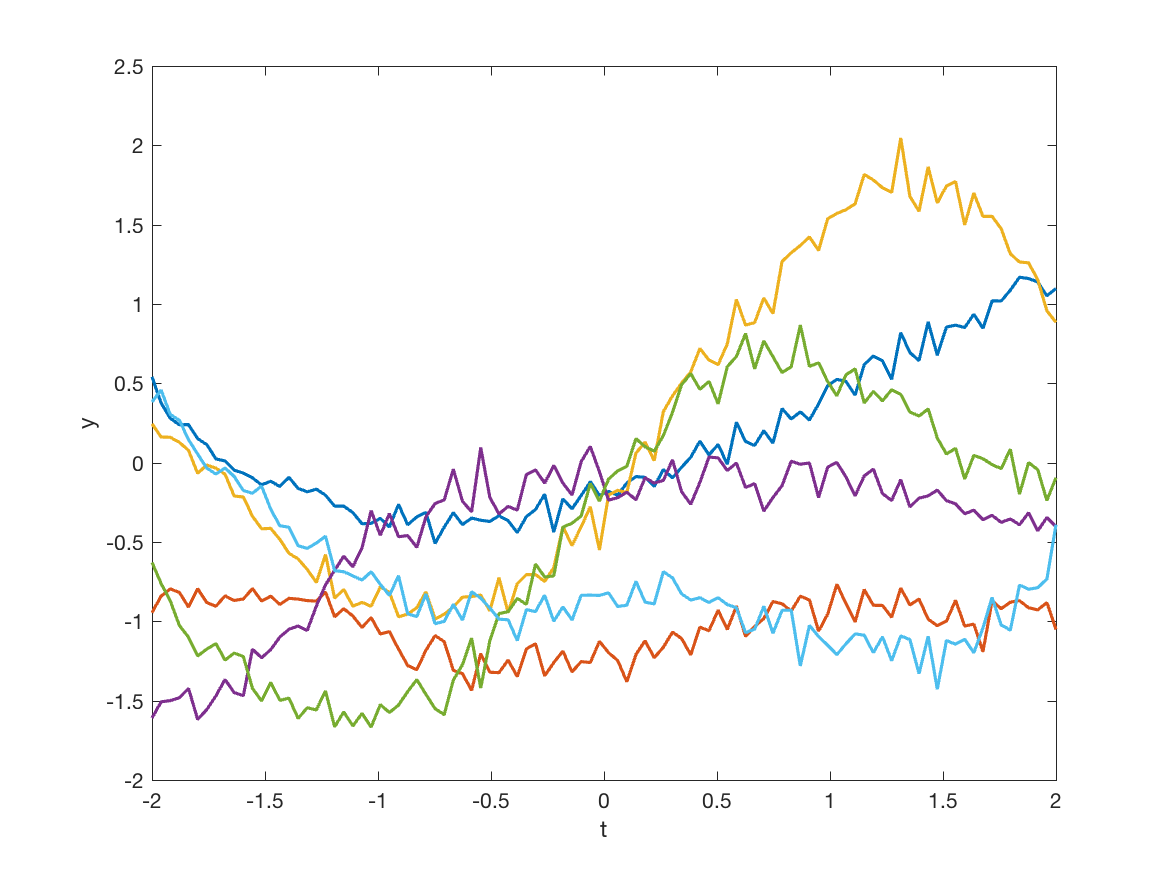}\label{fig:gp_sample_SE+WN}} 
	\caption{Samples from Gaussian processes with different covariance functions.}
\end{figure}

A key step in modeling time-series data with GPs, is then to define the mean and covariance functions. The mean function defines the mean of the process and, given adequate normalization of the observations, it is commonly taken to be a zero-value vector, i.e. $m(t) = 0$. As for the covariance function, it specifies basic aspects of the process, such as stationarity, isotropy, smoothness and periodicity. The most common choice of covariance function is the squared exponential (SE), which is defined as
\begin{align}
	k_{\mbox{\tiny SE}}(t,t') = h^2 \, \exp \bigg( - \frac{(t-t')^2}{2\ell^2} \bigg), 
\end{align}
with the parameter $\ell$ defining the characteristic length-scale and $h$ specifying an output-scale amplitude. Notice how the exponent goes to unity as $t$ becomes closer to $t'$. Hence, nearby points are more likely to covary. As a result, a GP prior with a squared exponential covariance function prefers smooth functions. Figure~\ref{fig:gp_sample_SE} shows 5 samples from this covariance function.  

Alternative popular choices of covariance functions for time-series include the Mat\'{e}rn, rational quadratic, white noise and periodic. All of these have different properties, and the choice of one over the other typically relies on knowledge from the domain. We describe here only the covariance functions used in this article and refer the interested reader to \cite{Rasmussen2005} for more details. Lastly, it is important to note that sums and products of covariance functions also produce valid covariance functions.

A particularly useful covariance function for time-series is the periodic (PER) covariance function, which is given by
\begin{align}
	k_{\mbox{\tiny PER}}(t,t') = h^2 \, \exp \bigg( - \frac{1}{2\ell^2} \sin^2 \bigg(\pi \frac{(t-t')^2}{p} \bigg) \bigg), 
\end{align}
where $h$ controls the amplitude and $p$ is the period. Another useful covariance function is the white noise (WN) covariance function with variance $\sigma^2$ defined as
\begin{align}
	k_{\mbox{\tiny WN}}(t,t') = \sigma^2 \, \delta(t,t'), 
\end{align}
where $\delta(t,t')$ is the Kronecker delta function which takes the value 1 when $t = t'$ and 0 otherwise. This covariance function allows us to account for uncertainty in the observed data and is typically found added to other covariance functions. Figures~\ref{fig:gp_sample_PER} to \ref{fig:gp_sample_SE+WN} show samples from these covariance functions.

Having specified a GP prior, $p(\bff) = \mathcal{GP}(m_f(t)=0, k_f(t,t'))$, for the function values $\bff$, the next step is to specify an appropriate likelihood function. If we are considering problems with continuous outputs, then perhaps the simplest and most common likelihood function to use is a Gaussian distribution with mean $f(t)$ and $\sigma^2$ variance. Letting $\by = \{y_t\}_{t=1}^T$ denote the observed time-series values, such that $y_t \in \mathbb{R}$, we have that $\by \sim \N(\by|\bff,\sigma^2\textbf{I}_T)$, where $\textbf{I}_T$ refers to the $T \times T$ identity matrix. Assuming a homoscedastic Gaussian likelihood of this form, the marginal distribution of $\by$ can be computed analytically as
\begin{align}
	p(\by) &= \int p(\by|\bff) \, p(\bff) \, d\bff  = \N(\by|\textbf{0}_T,\bV_f),
	\label{eq:marg_lik_exact}
\end{align}
where $\textbf{0}_T$ is used to denote a $T$-dimensional vector of zeros and $\bV_f \triangleq \sigma^2\textbf{I}_T + \textbf{K}_f$, with $\textbf{K}_f$ denoting the covariance function $k_f(t,t')$ evaluated between every pair of time indexes.

In time-series problems, our aim is often to make a prediction $y_*$ for an unobserved time $t_*$. The joint distribution over $y_*,y_1,...,y_T$ is simply given by
\begin{align}
	p(y_*,\by) = \N(y_*,\by|\textbf{0}_{T+1}, \bV_{f+1}), 
\end{align}
where, in order to keep the notation uncluttered, we omitted the (implicit) dependency on $\{t_*,\bt\}$, and we introduced
\begin{align}
	\bV_{f+1} \triangleq \left(\begin{array}{cc}\bV_f & \textbf{k}_{f*} \\\textbf{k}_{f*}^\transpose & k_{f**} + \sigma^2\end{array}\right). 
\end{align}
In the matrix above, we use $\textbf{k}_{f*}$ to denote the covariance function evaluated between the test point $t_*$ and all the other training points in $\bt$, and $k_{f**}$ to denote the covariance function evaluated between the test point $t_*$ against itself, i.e. $k_{f**} = k_f(t_*,t_*)$.

Using this joint distribution, we can now determine the distribution of $y_*$ conditioned on $\by$, \ie the predictive distribution, by making use of the conditional probability for Gaussians, yielding
\begin{align}
	p(y_*|\by) 
	&= \N(y_*|\textbf{k}_{f*}^\transpose \bV_f^{-1} \by, \, k_{f**} + \sigma^2 - \textbf{k}_{f*}^\transpose \bV_f^{-1} \textbf{k}_{f*}).
	\label{eq:pred_dist_exact}
\end{align}

One key advantage of the Bayesian formalism of GPs is the fully probabilistic interpretation of the predictions and their ability to handle uncertainty, which can be verified by plotting the predictive distributions in (\ref{eq:pred_dist_exact}) and noticing that the uncertainty is lower close to the observations and becomes higher as we go towards regions with no observations. In fact, besides providing us with a notion of confidence in the predictions, this Bayesian formalism also allows us to actively select which points to observe in situations where observations are costly to acquire, by choosing points where the uncertainty is highest. 

So far we have been assuming the hyper-parameters of the covariance function $k_f(t,t')$ to be fixed. However, these can be optimized by maximizing the logarithm of the marginal likelihood of the observations given in (\ref{eq:marg_lik_exact}).

\section{Sample-size-and-regime conditioned heteroscedastic GPs}
\label{sec:fc-hgp}

In the previous section, we assumed an uncorrelated zero-mean Gaussian distribution with a global or constant variance $\sigma^2$ for the noise term $\epsilon$ in (\ref{eq:homeskedastic_formulation}). This is the most common setting found in the time-series literature and it is referred to as the homoscedastic assumption. However, for many practical time-series problems of interest, this assumption is too restrictive and unrealistic. Heteroscedastic approaches relax this assumption by considering models of the form
\begin{align}
	y_t = f(t) + \epsilon_t,
	\label{eq:heteroscedastic_formulation}
\end{align}
where $\epsilon_t$ is a time-dependent noise term, such that $\epsilon_t \sim \N(\epsilon_t|0,r(t))$. This allows the uncertainty associated with each observation to vary with time. Notice how the homoscedastic setting corresponds to the special case when $r(t) = \sigma^2$. In order to ensure positivity, we parametrize $r(t) = e^{g(t)}$ and proceed by placing a GP prior on $g(t)$ as we did for $f(t)$, so that $g(t) \sim \mathcal{GP}(\mu_0,k_g(t,t'))$, where $\mu_0$ controls the scale of the noise process and $k_g(t,t')$ is the corresponding covariance function. 

The model described so far already allows us to account for time-varying uncertainty in the observations. Hence, if we consider, for example, a squared exponential covariance function for the noise process $g$, then we can already represent time-dependent observation noise that varies arbitrarily smoothly through time. However, crowdsourced traffic speeds that are obtained by aggregating GPS information of various sources (e.g., Google Maps users in the case of Google's traffic data) can vary significantly between consecutive time intervals due to various factors such as the number of GPS samples used to estimate the speed in a given road segment and their respective accuracy, traffic regime, etc. The same applies for other popular commercial traffic data providers, such as INRIX or HERE, who rely on GPS data from various sources ranging from users of their mobile phone application to contractual fleets (e.g., delivery vehicles and taxis). 

Fortunately, in gathering speed information from multiple GPS probes for a given road segment, crowdsourced traffic data providers also know the number of sample vehicles that traveled through that segment in a given time interval - the sample size. The latter can even be thought off as a proxy for the true (unobserved) traffic flow. However, it is important to stress that it does not correspond to the real flow, which would be unavailable for most probe vehicle traffic data providers. 

\begin{figure}
	\begin{center}
	\includegraphics[width=0.6\linewidth]{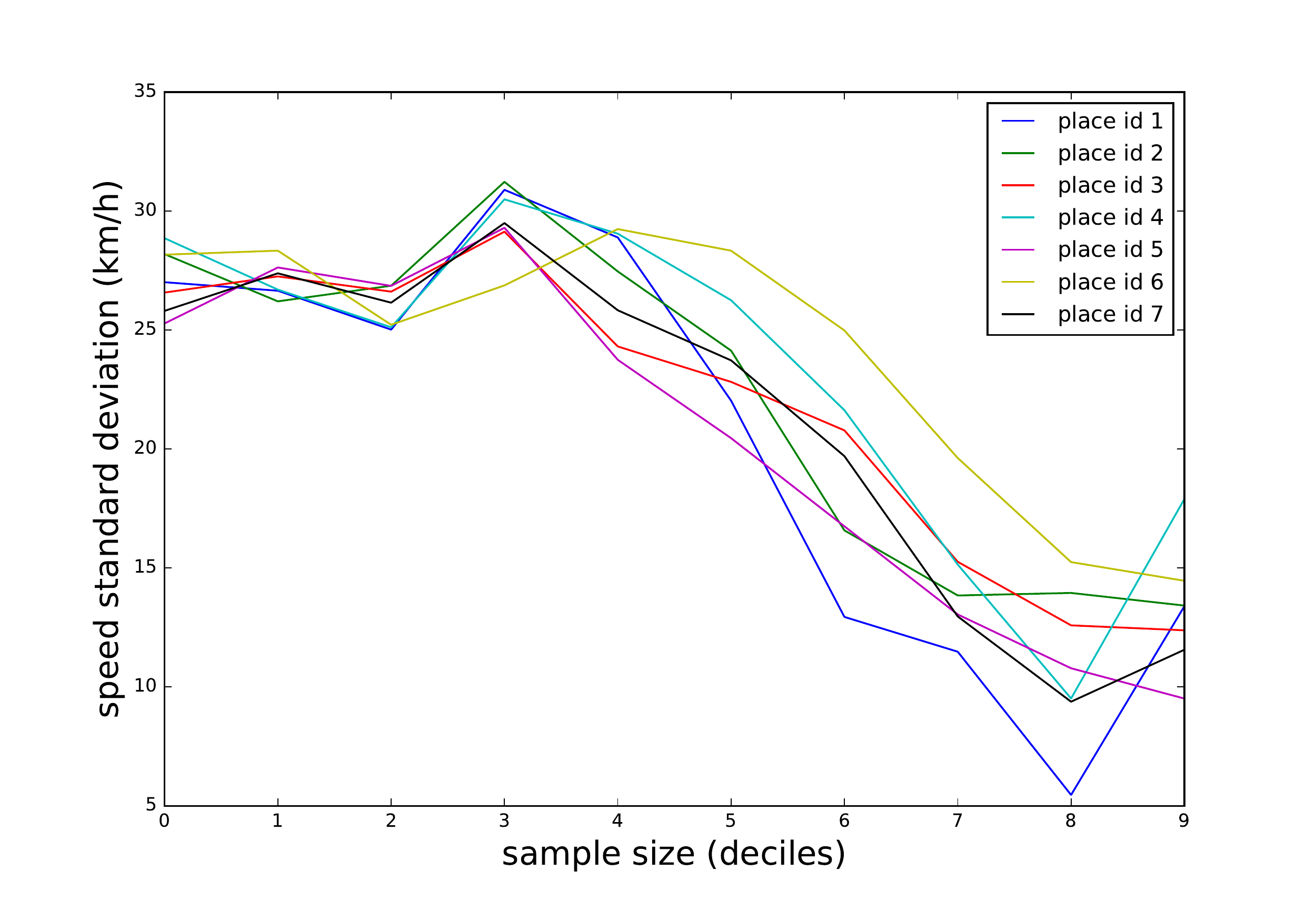} 
	\caption{Relationship between sample size and the standard deviation of the speeds at 7 road segments and over all 5-min intervals in the crowdsourced traffic data used in the experiments.}
	\label{fig:speed_std-flow}
	\end{center}
\end{figure}

Despite the fact that the observed sample size cannot be interpreted as true flow (but rather as a noisy proxy for it), we argue that there is a significative amount of valuable information in this data, especially for modeling the uncertainty in the observed speeds in a fully Bayesian framework, which we validate empirically in Section~\ref{sec:experiments}. Meanwhile, in order to try to provide some insights on the crowdsourced traffic data for Copenhagen used in the experiments and motivate the proposed approach, we provide a brief preliminary analysis. Figure~\ref{fig:speed_std-flow} shows the observed relationship between sample size and the standard deviation of speeds for 7 different random road segments according to our data, which suggests strong correlations between speed variance and sample size (provided by Google in deciles, i.e. discretized in 10 equally-sized bins) ranging from -0.78 to -0.88, with higher speed variances being associated with smaller sample sizes. However, the latter can be a consequence of two factors: (1) statistical effect resultant of the fact that variance tends to decrease with the size of the sample, or (2) a consequence of the D-shaped relationship between traffic speeds and flows as represented in the traffic fundamental diagrams (TFD), since low flows can correspond both to congestion and free flow conditions. 

\begin{figure}
	\begin{center}
	\includegraphics[width=0.9\linewidth,trim={2.7cm 2.cm 3.5cm 2.cm},clip]{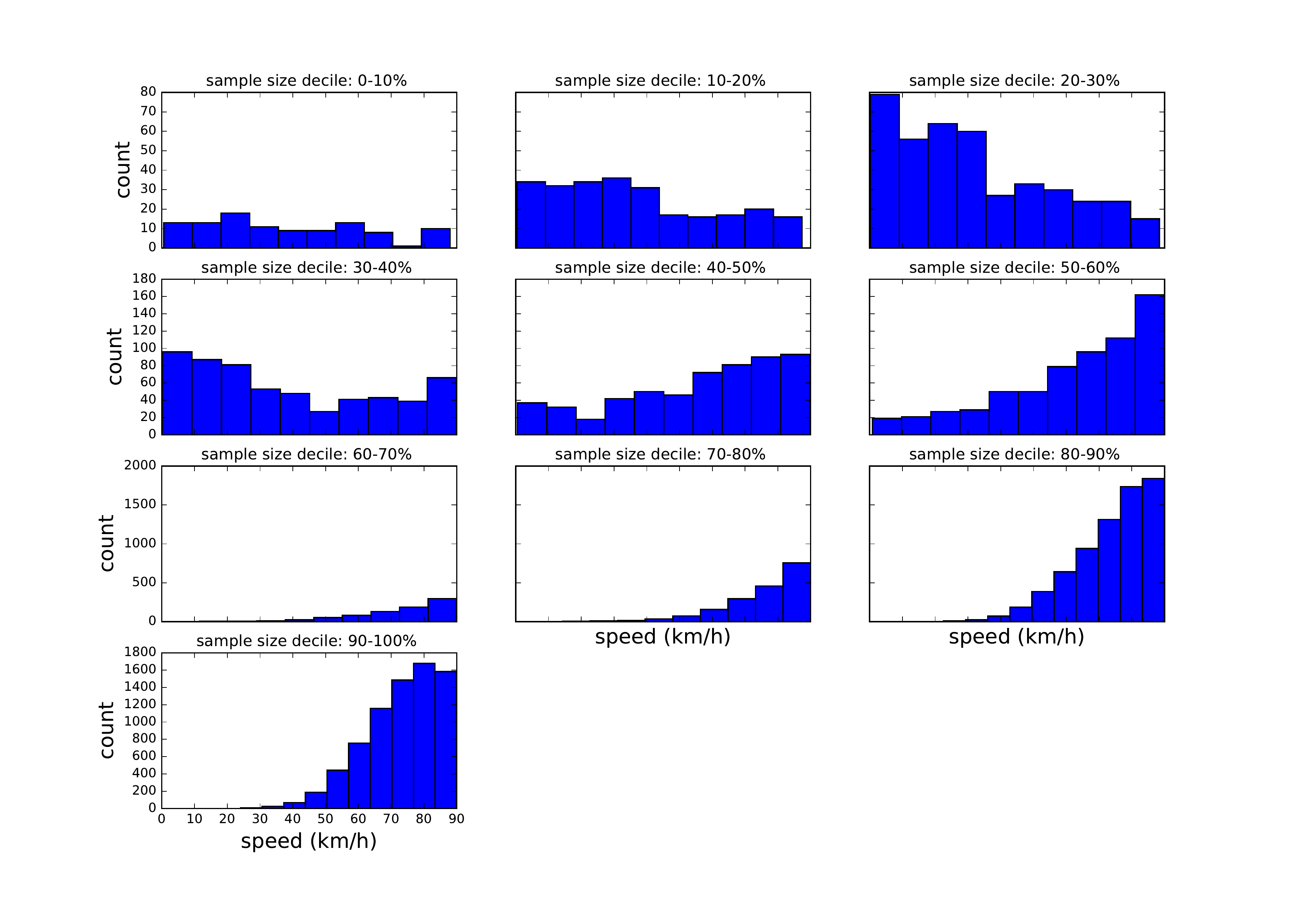} 
	\caption{Histograms of the speeds for each of the 10 sample size deciles (10 histograms) for an example road segment.}
	\label{fig:speed_dist_per_flow} 
	\end{center}
\end{figure}

In order to try to gain further insights, Figure~\ref{fig:speed_dist_per_flow} shows the distribution of speeds for the data corresponding to each sample size decile. In the figure, we can verify that low sample sizes have speeds across the entire speed domain and high sample sizes have speeds predominantly in the high speed domain. Under the assumption that the observed sample size is a good proxy for traffic flow, then this is consistent with the bi-valued relationship between speed and flow according to the TFD, but with one main difference: in this figure we have speeds in the middle, which is perfectly explained by the data coming from a mixed network. This emphasises the fact that the observation variance in the crowdsourced speed data can be dependent both on the sample size (as a statistical effect) and the traffic regime (congested vs. free-flow). 

\begin{figure}
	\begin{center}
	\includegraphics[width=0.9\linewidth,trim={2.7cm 2.cm 3.5cm 2.cm},clip]{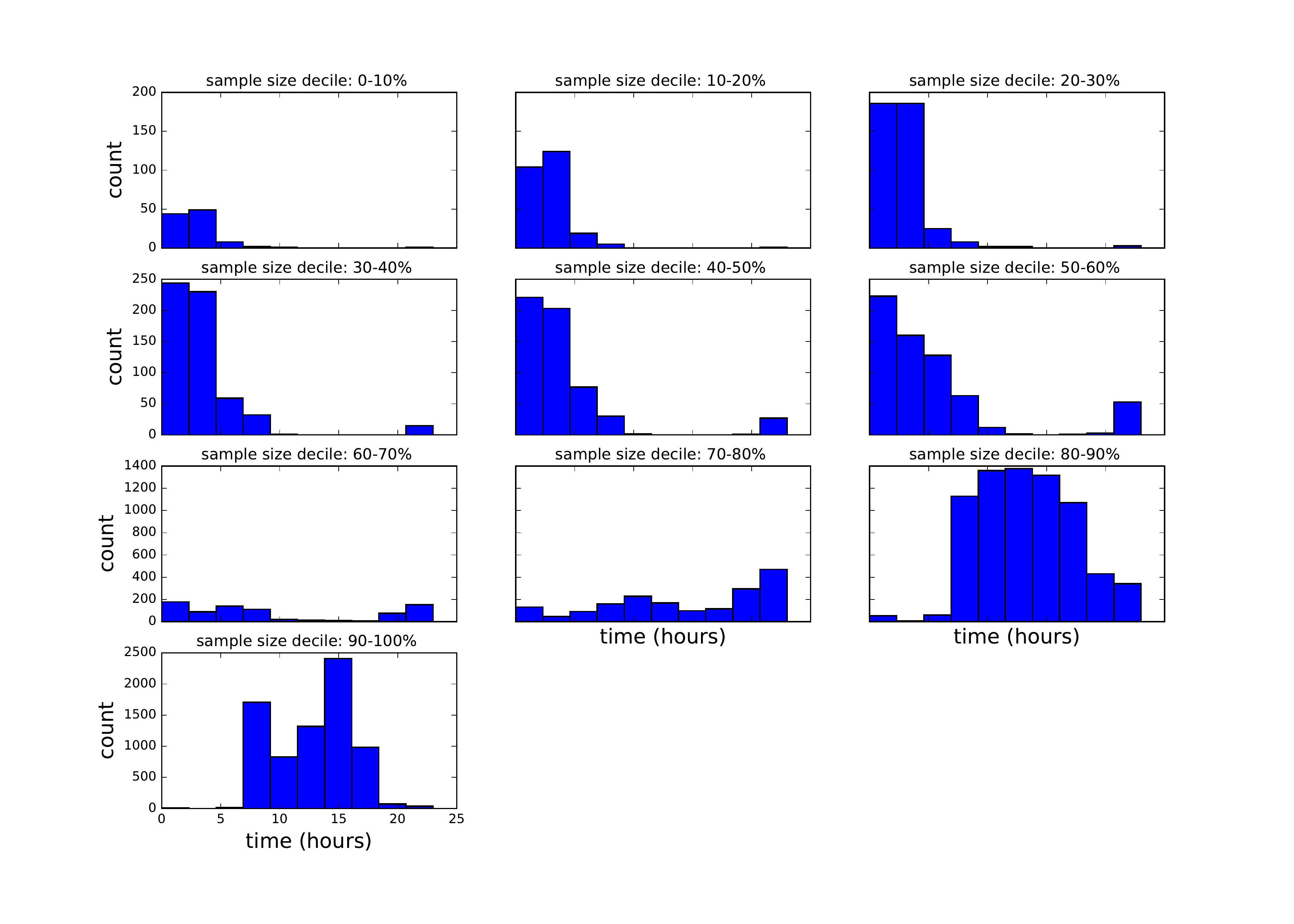} 
	\caption{Histograms of the hour of the day for each of the 10 sample size deciles (10 histograms) for an example road segment.}
	\label{fig:time_dist_per_flow}
	\end{center}
\end{figure}

Lastly, Figure~\ref{fig:time_dist_per_flow} shows the distribution of the different observed sample sizes among the different periods of the day. We can observe that the larger sample sizes tend to occur most frequently during day time. 


Exploiting all these insights, we propose a sample-size-and-regime-conditioned heteroscedastic Gaussian process (SSRC-HGP), where, instead of being dependent on time, the observation noise variance is dependent on the sample size and the traffic regime. Letting $x_t$ denote the observed sample size at time $t$, we would like to define the noise variance such that $r(x_t) = e^{g(x_t)}$. Unfortunately, when trying to predict the speed at time $t$, the sample size $x_t$ is also unavailable, i.e. it is a latent variable. We therefore proceed by modeling the sample size as a latent process indirectly via the Gaussian process for $g$. For this purpose, we shall take a \textit{function mapping} approach to time-series modeling \citep{roberts2013gaussian}, and model sample size without an explicit reference to the time ordering of the data. Instead, we assume the sample size at time $t$ to be a (non-linear) mapping from the $L$ previously observed sample sizes $\{x_{t-1},...,x_{t-L}\}$. 

Since low sample size can correspond both to high congestion and free flow conditions (bi-valued relationship between speed and flow), we further include a dependency of the observation variance on the previous observed speeds in the time-series, $\{y_{t-1},...,y_{t-L}\}$, as this allows the model to distinguishing between the two regimes purely from data. 
As such, we can define the noise variance as $r(x_{t-1},...,x_{t-L},y_{t-1},...,y_{t-L}) = e^{g(x_{t-1},...,x_{t-L},y_{t-1},...,y_{t-L})}$.

Let $\bgg = \{g(x_{t-1},...,x_{t-L},y_{t-1},..,y_{t-L})\}_{t=1}^T$, we can define a GP prior on $\bgg$ as\\ $p(\bgg) = \mathcal{GP}(\mu_0, k_g(\{x_{t-1},...,x_{t-L},y_{t-1},...,y_{t-L}\}, \{x'_{t-1},...,x'_{t-L},y'_{t-1},..,y'_{t-L}\}))$. The generative process of the proposed SSRC-HGP can then be summarized as:
\begin{enumerate}
	\itemsep0.1em
	\item Draw prior for mean function $\bff \sim \mathcal{GP}(0, k_f(t,t'))$
	\item Draw prior for noise function $\bgg \sim \mathcal{GP}(\mu_0, $\\ \mbox{ }\mbox{ }\mbox{ }\mbox{ }\mbox{ }\mbox{ }\mbox{ }\mbox{ }\mbox{ }\mbox{ }$k_g(\{x_{t-1},...,x_{t-L},y_{t-1},...,y_{t-L}\}, \{x'_{t-1},...,x'_{t-L},y'_{t-1},...,y'_{t-L}\}))$
	\item For each time $t \in \{1,\dots,T\}$
	\begin{enumerate}
		\item Draw speed $y_t \sim  \N \big(y_t|f(t),e^{g(x_{t-1},...,x_{t-L},y_{t-1},...,y_{t-L})} \big)$
	\end{enumerate}
\end{enumerate}

The marginal distribution of the observed speeds $\by$ is then given by
\begin{align}
	p(\by) &= \int p(\by|\bff,\bgg) \, p(\bff) \, p(\bgg) \, d\bff  \, d\bgg, 
\end{align}
where we omitted the dependency on $\bx$ in order to simplify the presentation. Unfortunately, this integral is no longer tractable and so is inference. Therefore, we proceed by using a variational inference algorithm similar to the one proposed in \cite{titsias2011variational} in order to perform approximate Bayesian inference in the proposed SSRC-HGP. The details are provided as supplementary material.\footnote{Supplementary material available at: \url{http://fprodrigues.com/supp-mat-fc-hgp.pdf}}

\section{Experiments}
\label{sec:experiments}

The heteroscedastic Gaussian process (HGP) approaches - standard HGP and our proposed SSRC-HGP\footnote{Source code available at: \url{http://www.fprodrigues.com/code_hgp_google.zip}} - were empirically evaluated using crowdsourced traffic data provided by Google, consisting of 6 months (January 2015 to June 2015) of traffic speeds and sample sizes for 7 road segments in Copenhagen that are known to be prone to traffic congestion. This dataset is derived from ``Location History" data that Google Maps users agreeingly share with Google. The individual GPS data is aggregated per road segment in 5 minute bins, resulting in a total of 51840 observations per road segment. The road segments are predefined by Google and uniquely identified using unique place IDs, whose details can be obtained through the use of the Google's Places API. Kindly notice how this traffic data is very similar to the one that is commonly found at public traffic agencies and local authorities, which is typically obtained by third-party commercial providers such as INRIX and HERE. 

For privacy and security reasons, the data does not contain the exact number of vehicles used to estimate the speed at each moment in a given segment (sample size). Instead, we are provided with an indicator of the sample size consisting of the distribution decile to which each observed number of samples belongs to. Therefore, if for a given road segment the sample size decile is 1, then we know that the number of samples (Google Maps users) traveling through that road segment in that 5 minute interval is in the bottom 10\% of the distribution (over the entire 6 months of data). Similarly, if the sample size decile is 8, then we know it is between the top 30\% and the top 20\% of the distribution of observed sample sizes. This is the information that will be provided to the proposed SSRC-HGP. Whenever speed or sample size information was not available in the original dataset, a simple linear interpolation method was used to infer the missing data.  


\begin{table}[t]
	\caption{Descriptive statistics of selected place IDs.}
	\small
	\begin{center}
		\begin{tabular}{c|c|c|>{\centering\arraybackslash}m{2cm}|>{\centering\arraybackslash}m{2cm}}
			Place ID & Mean speed & Speed std. & Mean sample size decile & Sample size decile std.\\
			\hline
			
			1 & 66.198 & 24.075 & 6.557 & 2.949\\
			2 & 76.849 & 18.525 & 7.622 & 1.775\\
			3 & 76.277 & 19.442 & 6.973 & 1.853\\
			4 & 66.276 & 26.451 & 6.357 & 2.770\\
			5 & 79.219 & 14.793 & 7.769 & 1.662\\
			6 & 62.375 & 27.015 & 5.783 & 1.922\\
			7 & 79.465 & 19.522 & 7.429 & 2.035
		\end{tabular}
	\end{center}
	\label{table:place_stats}
\end{table}%

\begin{figure}
	\begin{center}
	\includegraphics[width=0.6\linewidth]{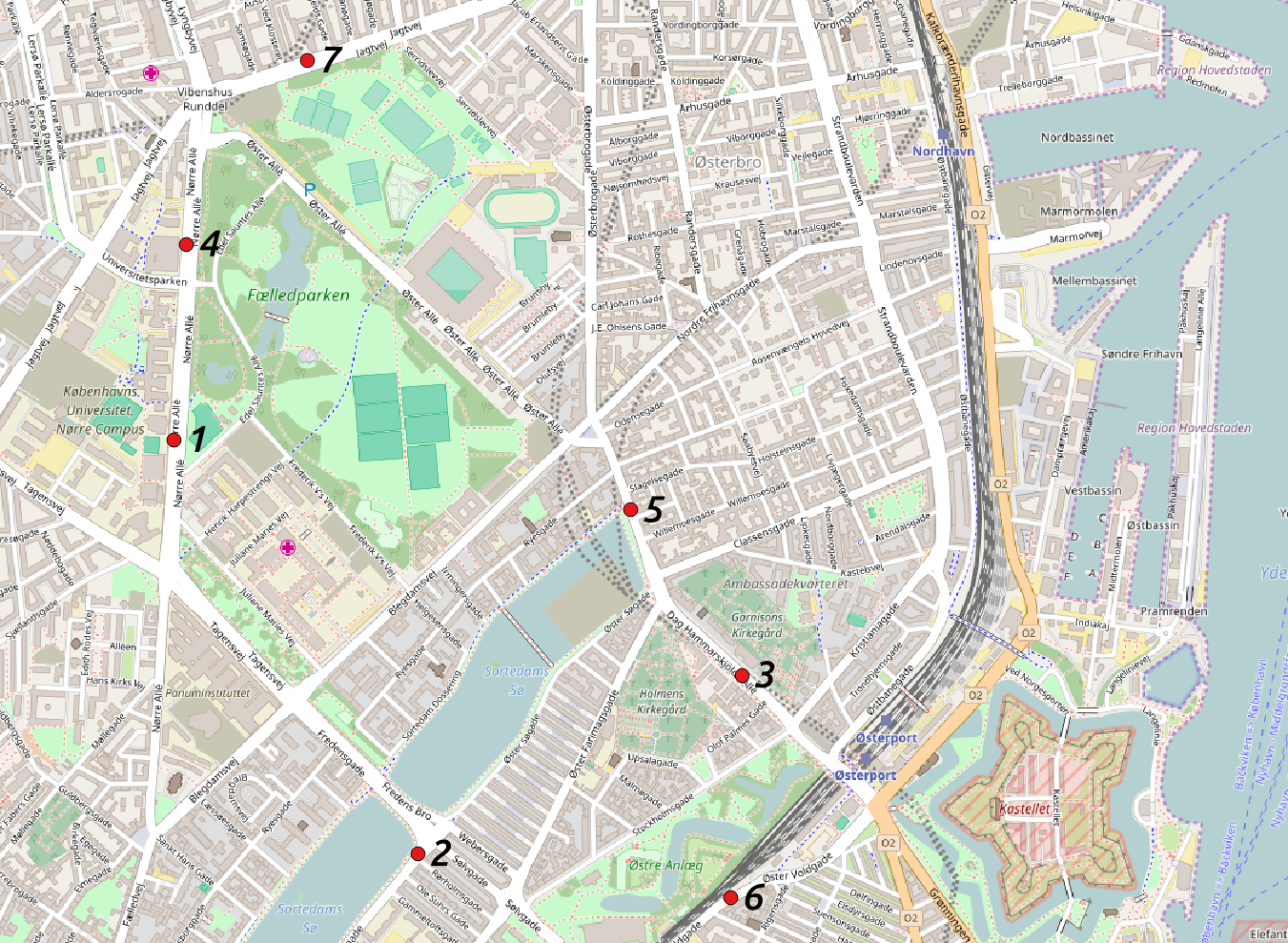} 
	\caption{Map of road segments used in the experiments.}
	\label{fig:selected_points}
	\end{center}
\end{figure}

Figure~\ref{fig:selected_points} shows the location of the selected road segments, and 
Table~\ref{table:place_stats} provides some descriptive statistics about these segments. As previously mentioned, for these 7 road segments, we consider two problems: speed imputation and short-term forecasting.

\subsection{Speed imputation}
\label{subsec:imputation}

In order to evaluate the performance of heteroscedastic GPs in imputation settings, we began by randomly selecting half of the observations for each road segment and removing them from the corresponding time-series. Special care was taken to ensure that the selected observations to be removed were not missing in the original dataset. The goal was then to try to infer the missing observations using the remaining information from the time-series. The approaches based on heteroscedastic GPs (HGP and SSRC-HGP) were compared with a standard (homoscedastic) Gaussian process (GP), which have already been shown to outperform traditional approaches such as ARIMA, SVM and KNN models for various transport-related tasks \citep{xie2010gaussian,neumann2009stacked,rodrigues2016bayesian}.  

For the homoscedastic GP, we used an additive covariance function that results from the combination of a squared exponential covariance function $k_{\mbox{\tiny SE}}(t,t')$ to model the smooth transitions in the speeds, a periodic covariance function $k_{\mbox{\tiny PER}}(t,t')$ to model repeatable weekly patterns, and a white noise covariance function $k_{\mbox{\tiny WN}}(t,t')$ to account for the observation noise. The covariance function used is then given by $k_f(t,t') = k_{\mbox{\tiny SE}}(t,t') + k_{\mbox{\tiny PER}}(t,t') + k_{\mbox{\tiny WN}}(t,t')$. The hyper-parameters of this covariance function were optimized by maximizing the marginal likelihood of the data. As for the heteroscedastic GP approaches, we used an additive covariance function for the mean speed process $f$ similar to one used in the homoscedastic case, but without the white noise component, since the HGP already accounts for observation noise by modeling it as a separate process. For the noise process $g$ in both HGP and SSRC-HGP, an additive covariance function was also used, consisting of a squared exponential and a white noise component. 

With the purpose of allowing for the parallelization of the different methods and to reduce the computational burden (particularly of the GPs), the 6 months of speed time-series for imputation were divided in smaller blocks and processed independently. We experimented with different block sizes and verified that above block sizes of 15 days the differences in the results were negligible. We therefore proceeded by considering blocks of 15 days for imputation, which allowed to greatly speed up the execution of the different methods without loss of imputation accuracy. This is particularly important when performing imputation in very large scale datasets, such as for the entire Copenhagen.   

As previously mentioned, in this article, we are mainly interested in the ability to model observation uncertainty and provide accurate prediction intervals for the predictions. In order to evaluate the quality of the predictive distributions provided by the different models, we use the negative log-predictive density (NLPD), which consists of the likelihood of the true speeds under the predictive distribution at test time indexes $t_*$, \ie $\mbox{NLPD} = -\log p(y_*|\by)$. Therefore, more accurate predictive distributions lead to lower values of NLPD. On the other hand, if the true speeds are in regions of low predictive probability density (\ie actual observed speed value is unlikely under the predictive distribution), then the value of NLPD will be high. For the standard (homoscedastic) GP, the NLPD can be readily evaluated by making use of (\ref{eq:pred_dist_exact}). However, for the HGPs we rely on the variational distribution $q(y_*)$ (kindly see supplementary material for details), which cannot be evaluated analytically. Instead, we approximate its value using Gauss-Hermite quadrature. 

\begin{table}[h!]
	\caption{Comparison of the accuracy of the predictive distribution produced by different imputation methods.}
	\small
	\begin{center}
		\begin{tabularx}{9cm}{@{} c | l | l | *{6}{>{\hsize=0.95\hsize}x} >{\hsize=1.35\hsize}x @{}}
			Place ID & Method & NLPD & ICP & MIL & RMIL\\
			\hline
			\multirow{3}{*}{\rotatebox[origin=c]{0}{1}} & GP &  0.627 & {0.954} & 1.830 & \mbox{ }85.935\\
			& HGP & 0.389 & 0.938 & 1.550 & 59.040\\
			& SSRC-HGP & \textbf{0.002} & 0.955 & {1.360} & {56.372}\\
			\hline
			\multirow{3}{*}{\rotatebox[origin=c]{0}{2}} & GP & 0.599 & 0.956 & 1.812 & 70.798\\
			& HGP & 0.348 & 0.934 & 1.413 & 66.111\\
			& SSRC-HGP & \textbf{0.068} & {0.950} & {1.372} & {37.098}\\
			\hline
			\multirow{3}{*}{\rotatebox[origin=c]{0}{3}} & GP &  0.771 & 0.953 & 2.045 & 84.278\\
			& HGP & 0.638 & 0.908 & {1.485} & {38.181}\\
			& SSRC-HGP & \textbf{0.325} & {0.944} & 1.577 & 58.536\\
			\hline
			\multirow{3}{*}{\rotatebox[origin=c]{0}{4}} & GP & 0.609 & 0.954 & 1.848 & 103.303\\
			& HGP & 0.520 & 0.912 & {1.317} & 71.167\\
			& SSRC-HGP & \textbf{0.034} & {0.946} & 1.355 & {40.402}\\
			\hline
			\multirow{3}{*}{\rotatebox[origin=c]{0}{5}} & GP & 0.602 & 0.952 & 1.874 & 65.338\\
			& HGP & 0.443 & 0.921 & {1.447} & 61.878\\
			& SSRC-HGP & \textbf{0.148} & {0.949} & 1.475 & {45.496}\\
			\hline
			\multirow{3}{*}{\rotatebox[origin=c]{0}{6}} & GP & 0.791 & {0.943} & 1.983 & 66.857\\
			& HGP & 1.276 & 0.845 & {1.417} & 50.561\\
			& SSRC-HGP & \textbf{0.570} & 0.923 & 1.625 & {48.069}\\
			\hline
			\multirow{3}{*}{\rotatebox[origin=c]{0}{7}} & GP & 0.740 & 0.954 & 2.085 & 127.126\\
			& HGP & 0.584 & 0.906 & 1.442 & 87.177\\
			& SSRC-HGP & \textbf{0.203} & {0.949} & {1.532} & {65.426}
		\end{tabularx}
	\end{center}
	\label{table:imput_bounds}
\end{table}%

Table~\ref{table:imput_bounds} shows the NLPDs obtained for the different methods that provide predictive distributions. As these results demonstrate, there is a general improvement by using HGPs instead of standard GPs, which is expected due to their capability to model the time-evolving observation uncertainty. However, when we analyze the results of the proposed SSRC-HGP, we can observe that there is a very substantial improvement in terms of the accuracy of the predictive distribution for all road segments considered. This shows that the information about the sample size is extremely valuable for modeling observation noise in the speeds. Indeed, this finding could be insightful to other areas beyond the transportation domain, where observation noise could be dependent on the sample size. 

At this point, it is important to emphasize that the log probability density of the test set observations is the most precise and reliable indicator of the quality of the predictive distribution and how well the different methods are modelings the uncertainty in the data. However, in an attempt to better understand some of the potential practical implications of these results, Table~\ref{table:imput_bounds} further shows metrics for evaluating the quality of the 95\% prediction intervals produced by the different methods. Contrarily to the predictive mean, the quality of the prediction intervals for the predictions is very difficult to evaluate and quantify, especially using a single criteria. Hence, our analysis of the 95\% prediction intervals will be based on the following statistics:
\begin{itemize}
	\item Interval coverage percentage (ICP), which corresponds to the fraction of the observations that are within the prediction intervals. Hence, for 95\% prediction intervals, this number should be close to $0.95$.
	\item Mean interval length (MIL), which measures the average length of the prediction intervals. 
	\item Relative mean interval length (RMIL) given by 
	\begin{align}
		\mbox{RMIL} = \frac{1}{T} \sum_{t=1}^T \frac{u_t-l_t}{|y_* - \hat{y}_*|}, 
	\end{align}
	where $u_t$ and $l_t$ refer to the upper and lower confidence bounds, respectively, and $\hat{y}_*$ is used to denote the predicted speed. RMIL intrinsically expresses the idea that, for a large observed error, we need to allow large intervals, so that the predicted value can be covered. 
\end{itemize}
Since the predictive distribution of the HGPs is non-Gaussian, we use Monte Carlo sampling in order to estimate confidence bounds. It is important to note that none of the statistics described above should be analyzed individually, but rather in the context of the others. For example, it is quite easy to produce prediction intervals with ICP close to 0.95 but with rather poor MIL and RMIL. Similarly, it is trivial to obtain confidence bounds with arbitrarily small MIL, but the ICP would also be lower than desired.  

From analyzing the statistics in Table~\ref{table:imput_bounds}, it becomes clear that the 95\% prediction intervals produced by the proposed SSRC-HGP are far more accurate than the other methods. In particular, while all methods seem to able to produce prediction intervals that cover roughly 95\% of the observations, the intervals produced by SSRC-HGP are much tighter (\ie lower MIL), especially when compared to standard GPs. The same applies to RMIL, thereby suggesting that the intervals produced by SSRC-HGP are larger in regions where the prediction error is expected to be large. It should be noted that, although for some road segments standard HGPs are able to produce tighter prediction intervals, they do so at the cost of interval coverage percentage (ICP), which in some cases can be as low as 84.5\%. 

\begin{figure}
	\centering
	\subfloat[][GP]{\includegraphics[trim={4cm 0 3.8cm 0},clip,scale=0.45]{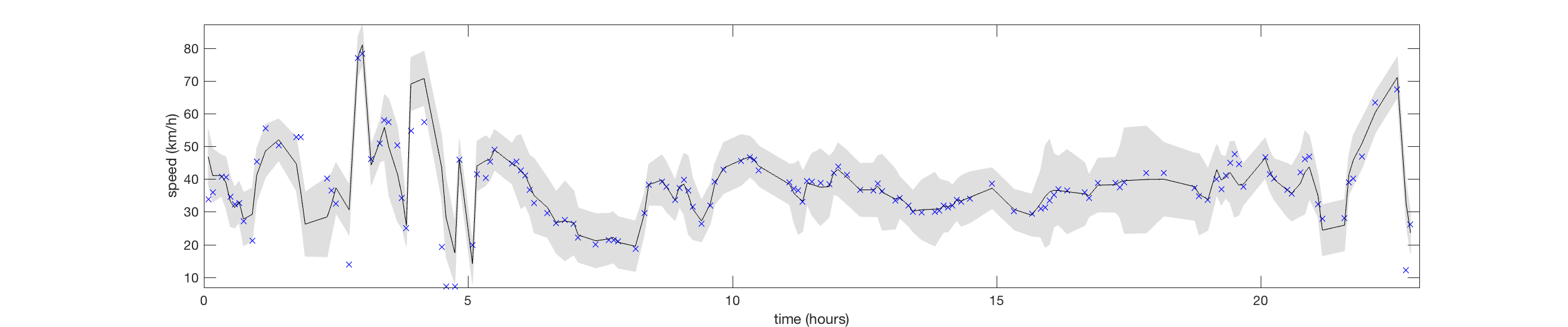}\label{fig:sample_imput_TA_gp}}
	
	\subfloat[][HGP]{\includegraphics[trim={4cm 0 3.8cm 0},clip,scale=0.45]{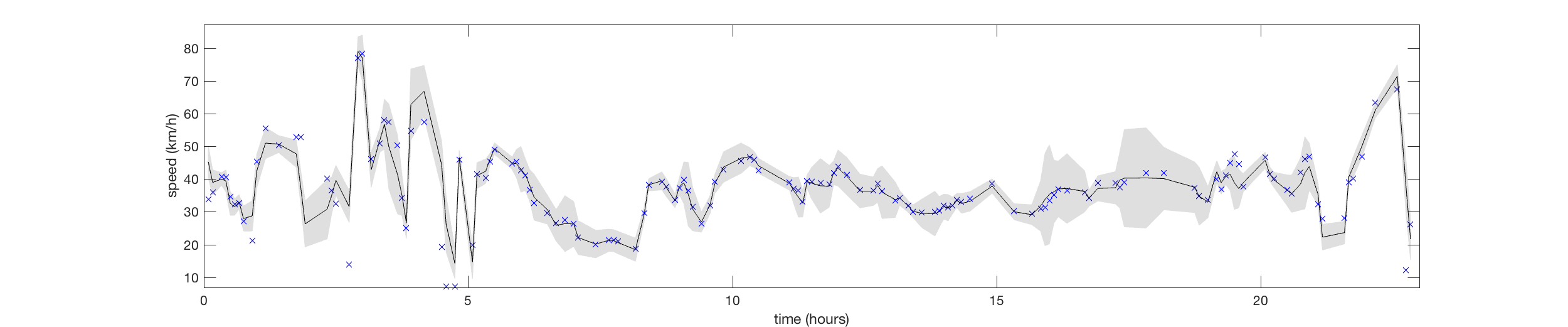}\label{fig:sample_imput_TA_hgp}}
	
	\subfloat[][SSRC-HGP]{\includegraphics[trim={4cm 0 3.8cm 0},clip,scale=0.45]{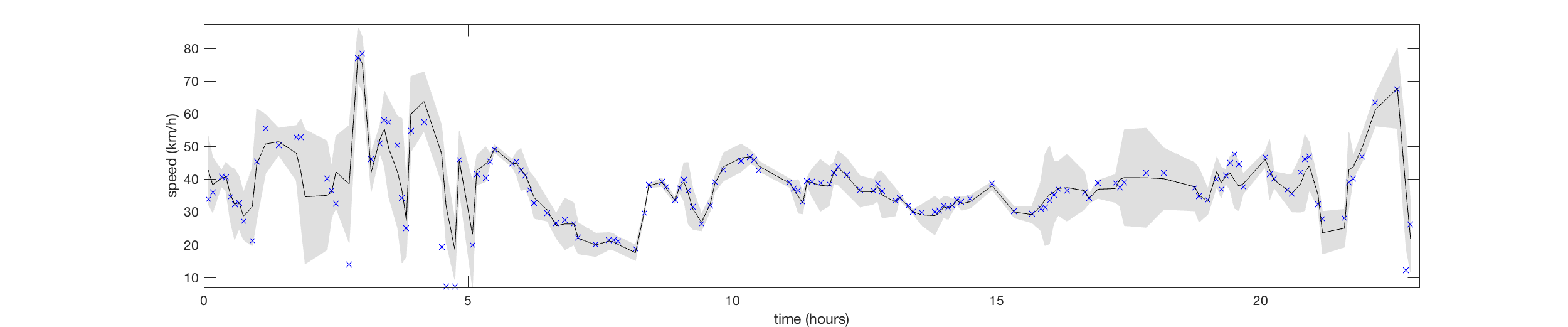}\label{fig:sample_imput_TA_hgp_flow}}
	\caption{Imputation results for the first 24 hours of speeds in place 1. Blue crosses represent the true speeds and the black lines with grey surrounding areas show the predicted speeds and corresponding 95\% prediction intervals.}
	\label{fig:samples_TA_imput}
\end{figure}

In order to gain even further insights, we plotted the prediction intervals produced by the different methods. Figure~\ref{fig:samples_TA_imput} shows the generated plots for the first 24 hours of predictions in place 1. We can see that the predictive distributions of the GP are rather uncertain and produce large prediction intervals. On the other hand, HGP is able to be much more confident in its predictions and produce much narrower prediction intervals, while still accounting for roughly 95\% of the data. 
However, at night times, when speed measurements from the mobile devices are not so reliable due to the low number of samples and other factors, the HGP tends to be overconfident in its predictions. By being conditioned on the sample size, the SSRC-HGP does not suffer from this issue and is capable of correctly modeling the uncertainty in the observed speeds during night periods. 

\begin{table}[h!]
	\caption{Accuracy of different imputation methods for the various road segments analyzed.}
	\small
	\begin{center}
		\begin{tabularx}{8cm}{@{} c | l | *{5}{>{\hsize=0.95\hsize}x} >{\hsize=1.35\hsize}x @{}}
			Place ID & Method & MAE & RAE & $R^2$\\
			\hline
			\multirow{4}{*}{\rotatebox[origin=c]{0}{1}} 
			& Lin. Interp. & 2.840 & 40.897 & 0.717\\
			& GP & 2.550 & 37.244 & 0.752 \\
			& HGP & 2.460 & 35.917 & 0.757 \\
			& SSRC-HGP & \textbf{2.414} & \textbf{34.764} & \textbf{0.759} \\ 
			\hline
			\multirow{4}{*}{\rotatebox[origin=c]{0}{2}} 
			& Lin. Interp. & 2.869 & 40.526 & 0.732\\
			& GP & 2.570 & 36.186 & 0.760\\
			& HGP & \textbf{2.440} & \textbf{34.361} & \textbf{0.775} \\
			& SSRC-HGP & 2.466 & 34.837 & 0.771\\ 
			\hline
			\multirow{4}{*}{\rotatebox[origin=c]{0}{3}} 
			& Lin. Interp. & 2.839 & 51.547 & 0.617\\
			& GP & 2.477 & 44.983 & 0.672 \\
			& HGP & 2.378 & 43.177 & 0.677 \\
			& SSRC-HGP & \textbf{2.336} & \textbf{42.412} & \textbf{0.694} \\ 
			\hline
			\multirow{4}{*}{\rotatebox[origin=c]{0}{4}} 
			& Lin. Interp. & 2.902 & 43.292 & 0.717\\
			& GP & 2.592 & 38.678 & 0.755 \\
			& HGP & {2.487} & {37.112} & {0.756} \\
			& SSRC-HGP & \textbf{2.464} & \textbf{36.766} & \textbf{0.757} \\ 
			\hline
			\multirow{4}{*}{\rotatebox[origin=c]{0}{5}} 
			& Lin. Interp. & 2.366 & 43.446 & 0.714\\
			& GP & 2.073 & 38.073 & 0.759 \\
			& HGP & \textbf{1.986} & \textbf{36.481} & \textbf{0.765} \\
			& SSRC-HGP & 2.020 & 37.112 & 0.755 \\ 
			\hline
			\multirow{4}{*}{\rotatebox[origin=c]{0}{6}} 
			& Lin. Interp. & 4.593 & 50.897 & 0.620\\
			& GP & 4.103 & 45.471 & 0.671 \\
			& HGP & 4.122 & 45.685 & 0.657 \\
			& SSRC-HGP & \textbf{3.945} & \textbf{43.724} & \textbf{0.690} \\ 
			\hline
			\multirow{4}{*}{\rotatebox[origin=c]{0}{7}} 
			& Lin. Interp. & 2.761 & 47.657 & 0.650\\
			& GP & 2.479 & 42.799 & 0.693 \\
			& HGP & {2.322} & {40.094} & {0.709} \\ 
			& SSRC-HGP & \textbf{2.232} & \textbf{39.258} & \textbf{0.714} \\
		\end{tabularx}
	\end{center}
	\label{table:imput_errors}
\end{table}%

Finally, we also analyzed the imputation accuracy of HGP and SSRC-HGP in comparison with standard GPs and also a linear interpolation method (``Lin. Interp."). Table~\ref{table:imput_errors} shows the obtained results. We report 3 evaluation metrics: mean absolute error (MAE), relative absolute error (RAE) and coefficient of determination ($R^2$). The results consistently show that the heteroscedastic approaches outperform all the baseline methods, especially in terms of MAE, where the improvements of HGPs over the best baseline method range from 3.9\% to 9.9\%.

\subsection{Real-time speed forecasting}

We will now turn to evaluating the heteroscedastic GP approaches on a speed forecasting task. The goal is thus to predict what the speed will be for $t+1$ given the previous observations up to time $t$, along with the uncertainty associated with the predictions. We compare the predictive distributions of HGP and SSRC-HGP with those of a (homoscedastic) GP and an ARIMA model\footnote{The order of the ARIMA model for each road segment was chosen using the Bayesian information criteria (BIC).}. All the GP-based approaches use the same additive covariance functions described in Section~\ref{subsec:imputation}. 

The experimental setup consists of a sliding window of size $N$ with increments of $1$ time step (5 minutes) in each iteration, where the $N$ last observations, $\{y_{t-N},\dots,y_{t}\}$, are used to predict the speed $y_{t+1}$. We experimented with different values for $N$ and found that for $N$ greater than 7-8 days, the predictive performance of the different approaches does not change. Therefore, in all experiments we consider $N = 8$ days. The first 15 days of the speeds time-series are used to bootstrap the different algorithms and we make predictions for the remaining five and a half months. Every 5 days, all statistical models are re-estimated, \ie the parameters of the ARIMA model are re-estimated though maximum likelihood and the hyper-parameters of the GP-based models are re-optimized by maximizing the marginal likelihood of the data.

\begin{table}[!htbp]
	\caption{Comparison of the accuracy of the predictive distribution produced by different forecasting methods.}
	\small
	\begin{center}
		\begin{tabularx}{12cm}{@{} c | l | l | l l l | l l l*{6}{>{\hsize=0.95\hsize}x} >{\hsize=1.35\hsize}x @{}}
			& & \multicolumn{4}{c|}{Eval: all periods} & \multicolumn{3}{c}{Eval: day periods}\\
			Place ID & Method & NLPD & ICP & MIL & RMIL & ICP & MIL & RMIL\\
			\hline
			\multirow{4}{*}{\rotatebox[origin=c]{0}{1}} & ARIMA & 0.796 & 0.949 & 18.257 & 53.787 & 0.994 & 18.264 & 65.089\\
			& GP & 0.679 & 0.961 & 18.193 & 52.555 & 0.996 & 18.213 & 60.978 \\
			& HGP & 0.453 & {0.950} & 14.987 & 48.268 & 0.995 & 14.506 & 55.857\\
			& SSRC-HGP & \textbf{0.049} & 0.955 & 11.062 & 32.179 & 0.987 & 8.644 & 35.123\\
			\hline
			\multirow{4}{*}{\rotatebox[origin=c]{0}{2}} & ARIMA & 0.674 & 0.959 & 17.761 & 50.249 & 0.995 & 17.760 & 53.118\\
			& GP & 0.653 & 0.962 & 17.909 & 43.940 & 0.995 & 17.897 & 51.152\\
			& HGP & 0.416 & {0.948} & 13.564 & 45.110 & 0.990 & 13.285 & 53.585\\
			& SSRC-HGP & \textbf{0.082} & 0.952 & 11.086 & 33.840 & 0.981 & 9.019 & 35.163\\
			\hline
			\multirow{4}{*}{\rotatebox[origin=c]{0}{3}} & ARIMA & 0.766 & {0.949} & 17.458 & 49.346 & 0.995 & 17.464 & 61.664\\
			& GP & 0.648 & 0.959 & 17.311 & 49.636 & 0.996 & 17.369 & 59.886\\
			& HGP & 0.421 & 0.953 & 14.535 & 49.692 & 0.995 & 14.208 & 59.244\\
			& SSRC-HGP & \textbf{0.025} & 0.955 & 10.755 & 34.508 & 0.987 & 8.381 & 39.421\\
			\hline
			\multirow{4}{*}{\rotatebox[origin=c]{0}{4}} & ARIMA & 0.783 & 0.942 & 18.017 & 50.174 & 0.994 & 18.023 & 59.777\\
			& GP & 0.728 & 0.960 & 18.114 & 51.081 & 0.993 & 20.263 & 48.937\\
			& HGP & 0.658 & 0.929 & 11.899 & 44.970 & 0.989 & 13.541 & 46.938\\
			& SSRC-HGP & \textbf{0.116} & 0.955 & 12.375 & 35.668 & 0.983 & 9.497 & 35.189\\
			\hline
			\multirow{4}{*}{\rotatebox[origin=c]{0}{5}} & ARIMA & 0.766 & 0.946 & 17.458 & 49.346 & 0.995 & 17.464 & 61.664\\
			& GP & 0.646 & 0.959 & 17.293 & 51.373 & 0.996 & 17.316 & 58.798\\
			& HGP & 0.418 & {0.953} & 14.514 & 49.029 & 0.995 & 14.255 & 60.422\\
			& SSRC-HGP & \textbf{0.030} & 0.955 & 10.778 & 35.714 & 0.987 & 8.409 & 39.845\\
			\hline
			\multirow{4}{*}{\rotatebox[origin=c]{0}{6}} & ARIMA & 0.766 & 0.945 & 17.458 & 49.346 & 0.995 & 17.464 & 61.664\\
			& GP & 0.649 & 0.959 & 17.344 & 51.734 & 0.996 & 17.596 & 64.397\\
			& HGP & 0.420 & 0.954 & 14.669 & 48.534 & 0.994 & 14.448 & 60.189\\
			& SSRC-HGP & \textbf{0.043} & 0.954 & 10.865 & 35.615 & 0.986 & 8.400 & 35.714\\
			\hline
			\multirow{4}{*}{\rotatebox[origin=c]{0}{7}} & ARIMA & 0.766 & 0.946 & 17.458 & 49.345 & 0.994 & 17.467 & 61.668\\
			& GP & 0.646 & 0.959 & 17.281 & 49.925 & 0.996 & 17.307 & 61.216\\
			& HGP & 0.428 & {0.953} & 14.715 & 50.679 & 0.994 & 14.180 & 60.831\\
			& SSRC-HGP & \textbf{0.040} & 0.951 & {10.890} & {33.974} & 0.987 & 8.508 & 38.165\\ 
		\end{tabularx}
	\end{center}
	\label{table:forecast_bounds}
\end{table}%

As before, and as with other works on heteroscedastic approaches \citep{kersting2007most,titsias2011variational}, in order to evaluate the quality of the predictive distributions produced by HGP and SSRC-HGP in comparison with those of the GP and ARIMA models, we use the negative log-predictive density (NLPD). Table~\ref{table:forecast_bounds} shows the obtained results, from which we can verify that the proposed SSRC-HGP is able to achieve remarkable results in terms of NLPD. This demonstrates that the predictive distributions produced by SSRC-HGP are significantly more accurate than both standard GP and HGP, which in turn implies that the SSRC-HGP is much better at modeling the time-varying uncertainty in crowdsourced traffic data. Not surprisingly, the results from Table~\ref{table:forecast_bounds} also show that Gaussian processes are able to produce better prediction intervals than ARIMA, thus confirming the ability of GPs to properly handle uncertainty, which stems from their fully Bayesian non-parametric formulation. In turn, the HGP is shown here to outperform standard GPs for all places studied, which is natural due to the fact that they explicitly model observation uncertainty as a function of time. However, what is truly interesting are the results obtained for the proposed SSRC-HGP, which are able to greatly improve over standard HGPs, with reductions in NLPD ranging from 84\% to 96\%.

As in the imputation setting, we also computed the interval coverage percentage (ICP), mean interval length (MIL) and relative mean interval length (RMIL) for the different methods. Although the NLPD is the most appropriate metric for comparing the accuracy of the predictive distributions produced by the different methods, since it accounts for the likelihood of the true speeds under the predictive distribution, we believe these other statistics give further insights on the results. However, the latter are restricted to evaluating a specific quantile of the predictive distribution, rather than the entire distribution as a whole. As in the imputation setting, we will focus on comparing the 95\% confidence bounds produced by the different methods.

As the results in Table~\ref{table:forecast_bounds} show, all methods are able to cover roughly 95\% of the observations ($\mbox{ICP} \approx 0.95$). However, the heteroscedastic approaches, especially the SSRC-HGP, are able to do so while producing significantly tighter prediction intervals, which can be verified by the results for the MIL. Furthermore, when we consider the relative version of MIL (RMIL), we can see that SSRC-HGP obtains much lower values, which suggests that the interval length is well adjusted according to the absolute error. 

\begin{figure}
	\centering
	\subfloat[][ARIMA]{\includegraphics[trim={4cm 0 3.8cm 0},clip,scale=0.45]{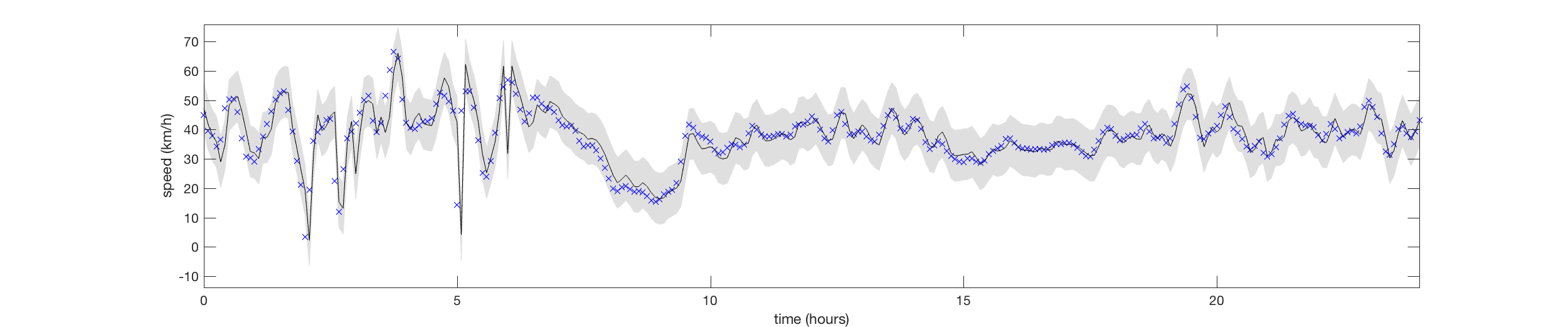}\label{fig:sample_Gc_arima}}
	
	\subfloat[][GP]{\includegraphics[trim={4cm 0 3.8cm 0},clip,scale=0.45]{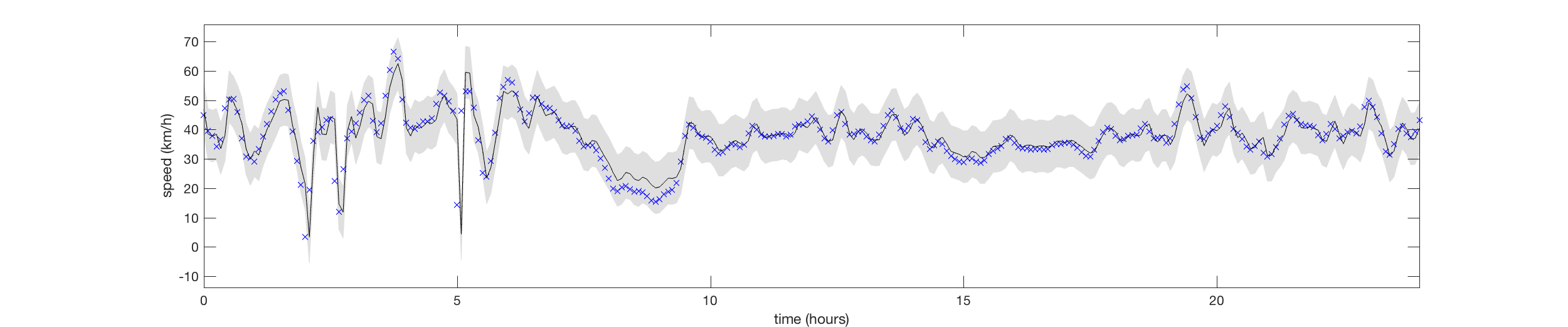}\label{fig:sample_Gc_gp}}
	
	\subfloat[][HGP]{\includegraphics[trim={4cm 0 3.8cm 0},clip,scale=0.45]{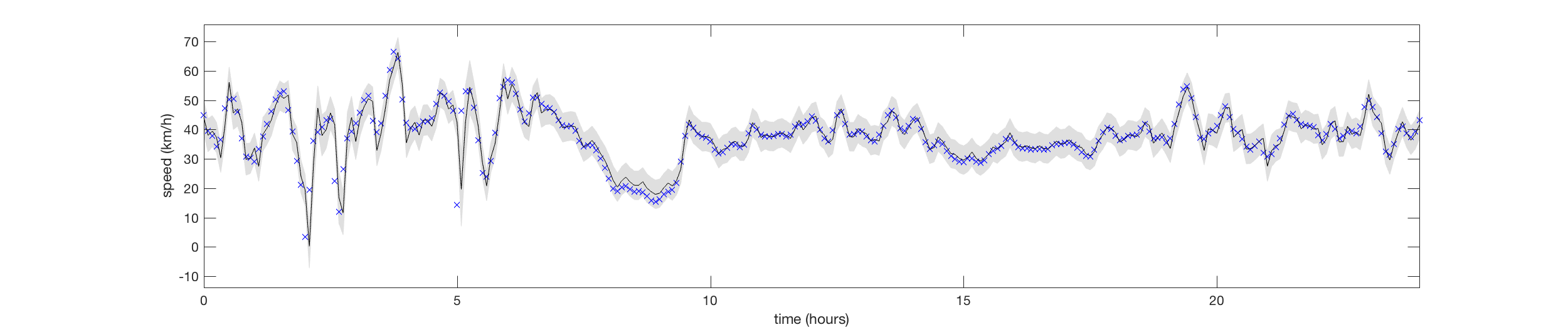}\label{fig:sample_Gc_hgp}}
	
	\subfloat[][SSRC-HGP]{\includegraphics[trim={4cm 0 3.8cm 0},clip,scale=0.45]{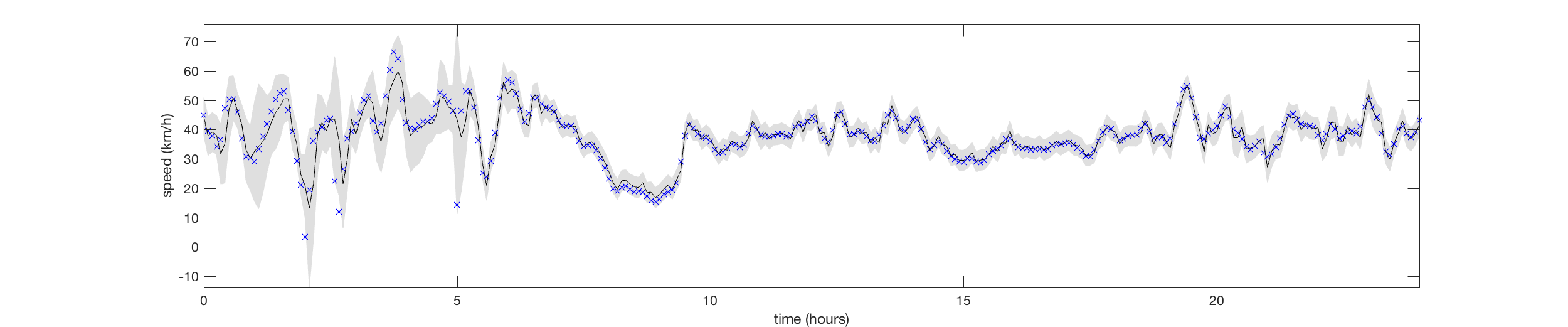}\label{fig:sample_Gc_hgp_flow}}
	\caption{Forecasting results for the first 24 hours of speeds in place 4. Blue crosses represent the true speeds and the black lines with grey surrounding areas show the predicted speeds and corresponding 95\% prediction intervals.}
	\label{fig:samples_Gc}
\end{figure}

In order to visually analyze the differences between the various statistical approaches, we plotted the predictions and their corresponding 95\% prediction intervals. Figure~\ref{fig:samples_Gc} shows the obtained plots for the first 24 hours in place 4, where the black lines correspond to the predicted speeds and the grey areas represent the prediction intervals. The blue crosses correspond to the observed true speeds. As Figure~\ref{fig:samples_Gc} shows, both ARIMA and GP produce nearly constant prediction intervals whose length is rather large. From a practical perspective, this could lead to poor optimization of services that are based on a specific confidence bound (\mbox{e.g.} the 97.5\% quantile of the distribution). 

On the other hand, we can see from Figure~\ref{fig:samples_Gc} that the prediction intervals of HGP are significantly tighter and much more accurate in general. However, as with the imputation case, we can verify that during night periods, the observed speeds vary significantly due mostly to the low number of samples of the Google data (\ie low number of mobile devices reporting speed data) and the standard HGP is not able to capture that uncertainty properly. Contrarily, by conditioning the uncertainty in speed on the latent sample size, the proposed SSRC-HGP is able to properly capture the uncertainty during night periods, thereby correctly learning that the prediction intervals should be larger during night and narrower during day periods or periods where the number of samples is higher. Moreover, we can notice that this ability of SSRC-HGP to correctly model observation uncertainty also leads to more accurate estimates of speed, which can be observed through the entire 24-hour sample and, in particular, by comparing the plots of the different methods during the morning peak period (between 7h00 and 9h30). In the night periods, we can see that its noise modeling capability prevents SSRC-HGP from predicting extreme values, which are generally unlikely. For example, around 5 a.m., we can see that, while all the other methods forecast a rather unlikely low value of speed, the proposed SSRC-HGP makes a much more moderate and sensible forecast but, at the same time, increases the prediction intervals in order to convey its uncertainty in that forecast.

%

\begin{table}[!htbp]
	\caption{Accuracy of different forecasting methods for the various road segments.}
	\small
	\begin{center}
		\begin{tabularx}{10.7cm}{@{} c | l | *{3}{>{\hsize=0.95\hsize}x} | *{3}{>{\hsize=0.95\hsize}x} >{\hsize=1.35\hsize}x @{}}
			& & \multicolumn{3}{c|}{Eval: all periods} & \multicolumn{3}{c}{Eval: day periods}\\
			ID & Method & MAE & RAE & $R^2$ & MAE & RAE & $R^2$\\
			\hline
			\multirow{4}{*}{\rotatebox[origin=c]{0}{1}} 
			& ARIMA & 2.845 & 40.830 & 0.718 & 1.780 & 33.467 & 0.878\\
			& GP & 2.639 & 38.180 & 0.765 & 1.767 & 33.210 & 0.884\\
			& HGP & 2.396 & 34.368 & 0.778 & 1.467 & 27.576 & 0.913\\
			& SSRC-HGP & \textbf{2.257} & \textbf{32.385} & \textbf{0.788} & \textbf{1.269} & \textbf{23.852} & \textbf{0.935}\\ 
			\hline
			\multirow{4}{*}{\rotatebox[origin=c]{0}{2}} 
			& ARIMA & 2.658 & 36.481 & 0.767 & 1.755 & 32.875 & 0.873\\
			& GP & 2.672 & 36.655 & 0.776 & 1.875 & 35.141 & 0.862\\
			& HGP & 2.343 & 32.142 & {0.802} & 1.523 & 28.537 & 0.899\\
			& SSRC-HGP & \textbf{2.271} & \textbf{31.174} & \textbf{0.807} & \textbf{1.413} & \textbf{26.461} & \textbf{0.909}\\ 
			\hline
			\multirow{4}{*}{\rotatebox[origin=c]{0}{3}} 
			& ARIMA & 2.748 & 39.560 & 0.732 & 1.712 & 32.823 & 0.885\\
			& GP & 2.565 & 36.870 & 0.778 & 1.663 & 31.869 & 0.894\\
			& HGP & 2.351 & 33.907 & {0.790} & 1.428 & 27.360 & 0.917\\
			& SSRC-HGP & \textbf{2.202} & \textbf{31.704} & \textbf{0.799} & \textbf{1.227} & \textbf{23.528} & \textbf{0.940}\\ 
			\hline
			\multirow{4}{*}{\rotatebox[origin=c]{0}{4}} 
			& ARIMA & 2.933 & 43.471 & 0.717 & 1.812 & 36.219 & 0.863\\
			& GP & 2.595 & 38.463 & 0.770 & 1.637 & 32.724 & 0.886\\
			& HGP & 2.294 & 34.008 & 0.770 & 1.283 & 25.636 & 0.921\\
			& SSRC-HGP & \textbf{2.128} & \textbf{33.299} & \textbf{0.788} & \textbf{1.199} & \textbf{23.821} & \textbf{0.935}\\ 
			\hline
			\multirow{4}{*}{\rotatebox[origin=c]{0}{5}} 
			& ARIMA & 2.748 & 39.560 & 0.732 & 1.712 & 32.823 & 0.885\\
			& GP & 2.555 & 36.764 & 0.779 & 1.662 & 31.829 & 0.895\\
			& HGP & 2.342 & 33.696 & {0.791} & 1.423 & 27.235 & 0.918\\
			& SSRC-HGP & \textbf{2.209} & \textbf{32.423} & \textbf{0.798} & \textbf{1.233} & \textbf{23.629} & \textbf{0.939}\\ 
			\hline
			\multirow{4}{*}{\rotatebox[origin=c]{0}{6}} 
			& ARIMA & 2.748 & 39.560 & 0.732 & 1.712 & 32.823 & 0.885\\
			& GP & 2.564 & 36.840 & 0.779 & 1.664 & 31.832 & 0.895\\
			& HGP & 2.359 & 33.886 & {0.789} & 1.430 & 27.360 & 0.917\\
			& SSRC-HGP & \textbf{2.201} & \textbf{31.692} & \textbf{0.798} & \textbf{1.224} & \textbf{23.468} & \textbf{0.940}\\ 
			\hline
			\multirow{4}{*}{\rotatebox[origin=c]{0}{7}} 
			& ARIMA & 2.748 & 39.560 & 0.732 & 1.712 & 32.823 & 0.885\\
			& GP & 2.563 & 36.943 & 0.779 & 1.669 & 32.055 & 0.893\\
			& HGP & 2.364 & 34.065 & 0.788 & 1.437 & 27.600 & 0.917\\
			& SSRC-HGP & \textbf{2.224} & \textbf{31.934} & \textbf{0.793} & \textbf{1.243} & \textbf{23.934} & \textbf{0.930} 
		\end{tabularx}
	\end{center}
	\label{table:forecast_errors}
\end{table}%

Lastly, we evaluate the performance of the different forecasting methods in terms of predictive mean. Table~\ref{table:forecast_errors} shows the results obtained. As these show, when we consider all periods of the day (``Eval: all periods") the heteroscedastic GP approaches clearly outperform all the other methods, with the sample-size-and-regime-conditioned HGP being the one that produces more accurate forecasts. Indeed, when compared to the best baseline method, SSRC-HGP leads to improvements in MAE between 12-18\%. Not surprising, the results also show that GPs outperform other standard methods such as ARIMA models, thereby confirming the findings of other works \citep{xie2010gaussian}. 

In order to further analyze the predictive accuracy of the different methods, Table~\ref{table:forecast_errors} also shows the same error statistics but evaluated only during day-time periods (``Eval: day periods"), \ie between 7 a.m. and 10 p.m. Interestingly, when we focus our evaluation on these periods, which are actually the periods that are most susceptible to congestion problems, the improvements of the heteroscedastic approaches over the baseline approaches become even more striking. For example, we can see that SSRC-HGP is able to exploit the sample size information to produce predictions that lead to reductions in MAE ranging from 25\% to 28\% when compared with standard GP, which are considered state of the art approaches for travel speed forecasting \citep{min2011real,ide2009travel}. Even when compared with the standard HGP, the proposed SSRC-HGP is able to reach improvements in MAE of 17\%. 

\section{Conclusion}
\label{sec:conclusion}

This article addressed the very fundamental problem in transport systems of traffic speed modeling for imputation of missing observations and real-time forecasting. Focusing on modern traffic data collection technologies based on crowdsourcing (\eg Google, INRIX, HERE, etc.), which are highly susceptible to fluctuations in the quality of the measurements, we proposed the use of heteroscedastic Gaussian processes (HGP) in order to model the non-constant observation noise. Furthermore, following the hypothesis that sample size is a key factor for explaining speed measurement noise, we proposed a sample-size-and-regime-conditioned heteroscedastic Gaussian process (SSRC-HGP), where the observation noise is conditioned on information about the sample size resultant from the GPS data collection process, as well as the traffic regime. 

We empirically showed, using a large-scale dataset of crowdsourced traffic speed data for the area of Copenhagen, that the heteroscedastic models outperform state-of-the-art approaches in terms of predictive accuracy and quality of the prediction intervals for imputation and real-time forecasting tasks. Particularly, in the case of the proposed SSRC-HGP, we verified that it can produce significantly more accurate predictive distributions in both tasks considered, which in turn lead to much more reliable prediction intervals. 
Given the practical importance of modeling observation noise and producing accurate prediction intervals when predicting different aspects of traffic and mobility in general, such as traffic speeds, flows or travel demand, we believe the ideas proposed in this article to constitute very important contributions. Moreover, it is important to note that these contributions go beyond the field of transportation systems, and extend to all applications of crowd-based remote-sensing where the sample size can be assumed to condition the observation noise. 

\section*{Acknowledgments}

The research leading to these results has received funding from the People Programme (Marie Curie Actions) of the European Union's Seventh Framework Programme (FP7/2007-2013) under REA grant agreement no. 609405 (COFUNDPostdocDTU), and from the European Union's Horizon 2020 research and innovation programme under the Marie Sklodowska-Curie Individual Fellowship H2020-MSCA-IF-2016, ID number 745673. 

The authors would also like to thank Google for proving access to the data used in this work.

\section*{References}

\bibliography{fc-hgp_v5}

\end{document}